\documentclass[a4paper,final]{article}
\usepackage{hyperref}		
\usepackage{listings}

\usepackage[utf8]{inputenc}
\usepackage[english,activeacute]{babel}
\usepackage{tabularx,wrapfig}
\usepackage{graphics}
\usepackage{graphicx}
\usepackage{epsfig}
\usepackage{amssymb,amsmath,amsthm,amsfonts}
\usepackage[all]{xy}
\usepackage{url}
\usepackage{subfigure}
\usepackage{chngpage}
\usepackage[figuresright]{rotating}
\usepackage{booktabs}
\usepackage{algorithm}
\usepackage{algorithmic}
\usepackage{comment}
\usepackage{morefloats}
\usepackage[shortlabels]{enumitem}

\newtheorem{definition}{Definition}

\begin{document}

\title{Induction of Decision Trees based on Generalized Graph Queries}
\author{Pedro Almagro-Blanco and Fernando Sancho-Caparrini}

\maketitle

\section{Introduction}

A decision tree is a classification (and regression) model that, based on the characteristics of a given object, and applying a series of rules, is able to classify it (or return a continuous value in the case of regression). The induction of decision trees from a set of previously classified objects is one of the most popular machine learning models due, among other things, to the low computational demand in their training and the interpretability of their results, so it is a representative white box model.

ID3 algorithm presented by R. Quinlan in 1983 for the automatic construction of decision trees from a training set of objects described through a collection of properties. Each object in the training set belongs to a \textit{class} (usually represented by the value of its \textit{target} attribute) of a set of mutually exclusive classes. ID3 has been one of the most used techniques in machine learning with applications to tasks as diverse as epidemics prediction, robot control, and automatic classification of clients in banks or insurance entities \cite{takale2004constructing, Salganicoff1996, He2012}.

The main goal of this work is to offer a methodology that allows to carry out machine learning tasks using decision trees on multi-relational graph data. In this context, the number of possible properties of each object goes far beyond those that it has directly associated, since the properties of the elements that are related to it can also be considered attributes of the object, and even the topological structure formed by the objects in their environment and the various measures that can be taken from the graph structure could be considered as additional attributes. With this objective, we will analyse different techniques that allow automatic induction of decision trees from graph data and we will present our proposal, GGQ-ID3, that aims to provide a framework to classify substructures in a graph, from simple nodes and edges, to larger paths and subgraphs, making use of Generalized Graph Query (GGQ) \cite{2017arXiv170803734A}.

This paper is structured as follows: we will start reviewing different techniques of induction of relational decision trees; then, we present our proposal based on the use of Generalized Graph Queries as evaluation tool; once our proposal is presented, we will show some examples of its application; and finally we present some conclusions and future work lines.

\section{Related Works}
\label{mrdtl}
 
This section does not pretend to be an exhaustive compilation of the related works that can be found in the literature, but a selection showing their predictive capacity, computational efficiency, or widespread use.

Inductive Logic Programming (ILP) \cite{plotkin1972automatic} is a machine learning area that uses Logic Programming to consistently represent examples, knowledge bases, and hypotheses. Although ILP itself (without proper transformation of the relationships between data to logical predicates) does not allow the generation of relational decision trees, it does allow automatic generation of logical decision trees that can be considered the basis of one of the most important relational decision tree algorithms, Multi-Relational Decision Tree Learning (MRDTL). ILP provides interpretability, but is inefficient when working with complex databases \cite{DBLP:journals/corr/abs-1211-3871}.

A \emph{Logical Decision Tree} \cite{ltrees} is a binary decision tree in which all the tests of the internal nodes are expressed as conjunction of literals of a prefixed First Order Language. The Top-Down Induction of Logical Decision Trees (TILDE) algorithm builds logical decision trees from a set of classified examples \cite{BLOCKEEL1998285}. The only difference between this algorithm and ID3 (without taking into account the possible optimizations implemented in C4.5 or others) is found in the tests carried out in each node of the tree. Following the rise of the ILP, some major breakthroughs were made in multi-relational data mining \cite{Kavurucu_confidence-basedconcept, 10.1109/GCIS.2009.289}. Yin Xiaoxin \cite{Yin2006} designed \emph{CrossMine}, a multi-relational classification model that merges ILP and relational databases, improving efficiency in this type of tasks through virtual joins \cite{improved}.

\emph{Multi-Relational Decision Tree Learning} (MRDTL) is a multi-relational machine learning algorithm \cite{Leiva02mrdtl:a} based on the ideas of Knobles et al. \cite{Knobbe99multi-relationaldecision} that works with Selection Graphs. A Selection Graph is a graph representation of a SQL query that selects records which match some constraints expressed by paths connected with them. In order to construct complex selection graphs from an initial one, some atomic operations allow to refine the queries. In addition, in order to be used as tests for ID3-like processes, it must be possible to obtain the complementary selection graph in each case. Essentially, MRDTL works as ID3, but making use of selection graphs as binary attributes in each decision node of the tree. The specification of this method is oriented to relational databases.

\emph{Graph-Based Induction} (GBI) is a data mining technique to perform network motifs mining from labelled and directed graphs through the union of pairs of connected nodes \cite{Nguyen2005}. It is very efficient because it uses a greedy technique. Tree Graph-Based Induction Decision (DT-GBI) is a decision tree construction algorithm to classify graphs using GBI principles. In DT-GBI the attributes (called patterns or substructures) are generated during the execution of the algorithm \cite{Geamsakul2003}, adding feature extraction capacity \cite{Nguyen2006}. It should be noted that, unlike our proposal or MRDTL, DT-GBI constructs decision trees to classify complete graphs, not general subgraphs (as is the case of GGQ-ID3) or nodes (as is the case of MRDTL).

\section{Generalized Graph Queries}
\label{GGQ}

Generalized Graph Query, on which the multi-relational decision tree induction model GGQ-ID3 is based, is briefly presented now. A more comprehensive description of this framework can be found in \cite{2017arXiv170803734A}.

A \emph{Generalized Graph} (which will also be called \emph{Property Graph}) is a concept that covers several variants of graphs that can be found in the literature.

\begin{definition}
A \emph{Generalized Graph} is a tuple $ G = (V, E, \mu) $ where:
\begin{itemize}
\item $ V $ and $ E $ are sets, called, respectively, \emph{set of nodes} and \emph{set of edges} of $ G $.
\item $ \mu $ is a relation (usually functional, but not necessarily) that associates each node or edge in the graph with a set of properties, that is, $ \mu: (V \cup E) \times R \rightarrow S $, where $ R $ represents the set of possible \emph{keys} for available properties, and $ S $ the set of possible \emph{values} associated.
\end{itemize} 

Usually, for each $\alpha \in R$ and $x\in V\cup E$, we write $ \alpha (x) = \mu (x,\alpha) $.

In addition, we require the existence of a special key for the edges of the graph, called \emph{incidences} and denote by $ \gamma $, which associates to each edge of the graph a tuple, ordered or not, of vertices of the graph.
\end{definition}

Generalized Graph Query allows structural and semantic, accurate, optimal, and based on a type of Regular Pattern Matching queries, associating edges of the pattern with paths on the graph that meet the constraints imposed, and allowing cycles.

Let $ L $ be a First Order Language with equality using a collection containing all the functions of $ \mu $ and constants associated with each element of the graph as non logical symbols (and some additional symbols, for example metrics defined on the elements of the graph). Constructing in the usual way the set of terms of the language and the set of formulas, $ FORM (L) $ (which we will call predicates), we can define queries on generalized graphs:

\begin{definition}
    A \emph{Generalized Graph Query (GGQ)} over $L$ is a binary generalized graph, $Q = (V_Q,E_Q,\mu_Q)$, where exist $\alpha$ and $\theta$, properties in $\mu_Q$, such that:
    \begin{itemize}
        \item $\alpha:V_Q\cup E_Q\rightarrow \{+,-\}$ total.
        \item $\theta:V_Q\cup E_Q\rightarrow FORM(L)$ associates a binary predicate, $\theta_x$, to every element $x$ of $V_Q\cup E_Q$.
    \end{itemize}
\end{definition}

The second parameter of $\theta$ is used to express conditions of membership on subgraphs and the first one will receive elements of the corresponding type to which it is associated (nodes or edges/paths).

Given a GGQ, $ x^+ $, respectively $ x^- $, will indicate that $ \alpha (x) = + $, respectively $ \alpha (x) = - $, and $V_Q^+/V_Q^-$ (respectively, $E_Q^+/E_Q^-$) the set of positive/negative nodes (respectively, edges). If for an element, $ \theta_x $ is not explicitly defined, we assume it to be a tautology (generally denoted by $T$). As we will see below, positive elements of the pattern represent elements verifying the associated predicates that must be present in the graph, while negative ones represent elements that should not be present in the graph.

In order to express the necessary conditions that define the application of a GGQ on a graph, the following notations are useful:

\begin{definition}
    Given a GGQ, $Q=(V_Q,E_Q,\mu_Q)$, the set of $Q$-predicates associated to $Q$ is:
    \begin{enumerate}
        \item For each edge, $e\in E_Q$, we define:
        $$Q_{e^o}(v,S)=\exists \rho\in \mathcal{P}_v^o(G)\ \left(\theta_e(\rho,S)\wedge \theta_{e^o}(\rho^o,S) \wedge \theta_{e^i}(\rho^i,S)\right)$$
        $$Q_{e^i}(v,S)=\exists \rho\in \mathcal{P}_v^i(G)\ \left(\theta_e(\rho,S)\wedge \theta_{e^o}(\rho^o,S) \wedge \theta_{e^i}(\rho^i,S)\right)$$

        In general, we will write $Q_{e^*}(v,S)$, where $*\in \{o,i\}$, and we will denote:
        $$Q_{e^*}^+ = Q_{e^*},\hspace{1cm} Q_{e^*}^- = \neg Q_{e^*}$$
        \item For each node, $n\in V_Q$, we define:
        \begin{align*}
        Q_n(S)&=\exists v\in V\ \left(\bigwedge_{e\in \gamma^o(n)} Q_{e^o}^{\alpha(e)}(v,S)\ \wedge \bigwedge_{e\in \gamma^i(n)} Q_{e^i}^{\alpha(e)}(v,S)\right)\\
        & = \exists v\in V\ \left(\bigwedge_{e\in \gamma^*(n)} Q_{e^*}^{\alpha(e)}(v,S)\right)
        \end{align*}
        That can be written generally as:
        $$Q_n(S)=\exists v\in V\ \left(\bigwedge_{e\in \gamma(n)} Q_{e}^{\alpha(e)}(v,S)\right)$$
        In addition, we denote:
        $$Q_n^+ = Q_n,\hspace{1cm} Q_n^- = \neg Q_n$$
    \end{enumerate}
\end{definition}

Where $ e ^ o / e ^ i $ represents the source/target node of the edge $ e $, $ \mathcal{P}_v^o (G)/ \mathcal{P}_v^i(G) $ represents the paths in $ G $ starting/ending  in node $ v $.

From these notations, we can formally define when a subgraph matches a given GGQ:

\begin{definition}
\label{verifica}
    Given a subgraph $S$ of a property graph, $G=(V,E,\mu)$, and a Generalized Graph Query, $Q=(V_Q,E_Q,\mu_Q)$, both over a language $L$, we say that $S$ \emph{matches} $Q$, $S\vDash Q$, if it is verified:

    $$Q(S)=\bigwedge_{n\in V_Q} Q_n^{\alpha(n)}(S)$$

    Otherwise, we will write: $S\nvDash Q$.
\end{definition}

In order to obtain controlled query generation methods, refinements can be defined in order to modify a GGQ by unit steps. Given two GGQ, $ Q_1, \ Q_2 $, we say that $ Q_1 $ \emph{refine} $ Q_2 $ in a graph $ G $, $ Q_1 \preceq_G Q_2 $, if for all $ S \subseteq G $ that $ S \vDash Q_1 $, then $ S \vDash Q_2 $.

\begin{definition}
    Given $Q\in GGQ$, $R\subseteq GGQ$ is a \emph{refinement set} of $Q$ in $G$ if:
    \begin{enumerate}
        \item $\forall\ Q'\in R\ (Q'\preceq_G Q)$
        \item $\forall\ S\subseteq G\ (S\vDash Q\Rightarrow \exists !\ Q'\in R\ (S\vDash Q'))$
    \end{enumerate}
\end{definition}

In what follows, given $ Q $, $Cl_Q^W$ is the graph obtained from $ Q $ by duplicating the nodes in $ W \subseteq V_Q $ (with the incident edges if they have).

Following \cite{2017arXiv170803734A}, we can prove that the following sets of GGQ are refinements of $ Q $:

\begin{itemize}
	\item \textbf{Add new node}: if $m\notin V_Q$, then $Q+\{m\}$:
        \begin{align*}
        Q_1 &= (V_Q\cup\{m\},\ E_Q,\ \alpha_Q\cup(m,+),\ \theta_Q\cup(m,T))\\
        Q_2 &= (V_Q\cup\{m\},\ E_Q,\ \alpha_Q\cup(m,-),\ \theta_Q\cup(m,T))
        \end{align*}
	\item \textbf{Add new edge between positive nodes}: if $n,m\in V_Q^+$, then $Q+\{n^+\overset {e^*}{\longrightarrow} m^+\}$ ($*\in\{+,-\}$) (where $Q'=Cl_Q^{\{n,m\}}$):
		\begin{align*}
		Q_1 &= (V_{Q'},\ E_{Q'}\cup\{n^+\overset {e^*}{\longrightarrow} m^+\},\ \theta_{Q'}\cup(e,T))\\
		Q_2 &= (V_{Q'},\ E_{Q'}\cup\{n^+\overset {e^*}{\longrightarrow} m^-\},\ \theta_{Q'}\cup(e,T))\\
		Q_3 &= (V_{Q'},\ E_{Q'}\cup\{n^-\overset {e^*}{\longrightarrow} m^+\},\ \theta_{Q'}\cup(e,T))\\
		Q_4 &= (V_{Q'},\ E_{Q'}\cup\{n^-\overset {e^*}{\longrightarrow} m^-\},\ \theta_{Q'}\cup(e,T))
		\end{align*}
	\item \textbf{Add predicate to positive edge connecting positive nodes}: if $n,m\in V_Q^+$, with $n^+\overset {e^+}{\longrightarrow} m^+$, and $\varphi\in FORM$, then $Q+\{n^+\overset {e \wedge \varphi}{\longrightarrow} m^+\}$ (where $Q'=Cl_Q^{\{n,m\}}$):
		\begin{align*}
		Q_1 &= (V_{Q'},\ E_{Q'}\cup\{n^+\overset {e'}{\longrightarrow} m^+\},\ \theta_{Q'}\cup(e',\theta_e\wedge \varphi))\\
		Q_2 &= (V_{Q'},\ E_{Q'}\cup\{n^+\overset {e'}{\longrightarrow} m^-\},\ \theta_{Q'}\cup(e',\theta_e\wedge \varphi))\\
		Q_3 &= (V_{Q'},\ E_{Q'}\cup\{n^-\overset {e'}{\longrightarrow} m^+\},\ \theta_{Q'}\cup(e',\theta_e\wedge \varphi))\\
		Q_4 &= (V_{Q'},\ E_{Q'}\cup\{n^-\overset {e'}{\longrightarrow} m^-\},\ \theta_{Q'}\cup(e',\theta_e\wedge \varphi))
		\end{align*}
	\item \textbf{Add predicate to positive node with positive environment}: if $n\in V_Q^+$, $\mathcal{N}_Q(n)\subseteq V_Q^+$, and $\varphi\in FORM$, then $Q+\{n\wedge \varphi\}$:
		$$\{Q_{\sigma}=(V_{Q'},E_{Q'},\alpha_{Q'}\cup \sigma,\theta_{Q'}\cup(n',\theta_n\wedge\varphi))\ :\ \sigma\in \{+,-\}^{\mathcal{N}_Q(n)}\}$$
	where $Q'=Cl_Q^{\mathcal{N}_Q(n)}$, and $\{+,-\}^{\mathcal{N}_Q(n)}$ is the set of all possible sign assignations to the elements in $\mathcal{N}_Q(n)$ (the neighborhood of $n$ in $Q$).
\end{itemize}

\begin{figure}[h]
    \begin{center}
        \includegraphics[scale=0.25]{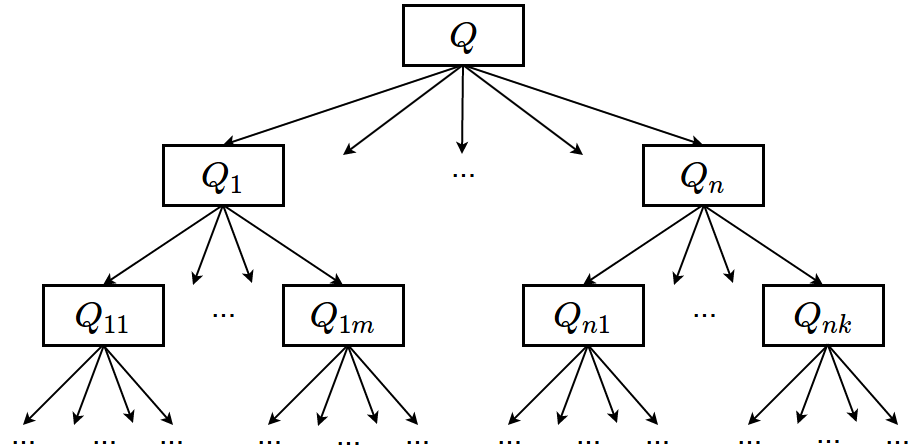}
    \end{center}
    \caption{%
        Refinement tree
    }%
    \label{arbolGGQ}
\end{figure}

It is an open problem how to automatically obtain a complementary GGQ of a given one. Refinements  cover this gap and allow, for example, the construction of an embedded partitions tree with nodes labelled as follows (Fig. \ref{arbolGGQ}):

\begin{itemize}
    \item The root node is labelled $ Q_0 $ (any initial GGQ).
    \item If a tree node is labelled $ Q $, and $ R = (Q_1, \dots, Q_n) $ is a refinement set of $ Q $, then its child nodes are labelled with the $ R $ elements.
\end{itemize}

Note that the construction of this tree completely depends on the refinement chosen in each branch and the initial GGQ.

\section{GGQ-ID3}

GGQ-ID3 is an adaptation of ID3 algorithm to create decision trees to classify structures immersed in a property graph using Generalized Graph Query framework as test tools in each decision node of the decision tree.

Similarly to ID3, our proposal, using refinements, will look for GGQs that best classify the examples of the training set along the tree construction (feature extraction). For that, in each internal node of the tree a discrimination will be performed between the structures of a graph $G$ that fulfil each GGQ resulting from a refinement. As usual, the measure of impurity used is a hyper-parameter of the algorithm.

Although GGQ-ID3 has a very similar structure to other algorithms for construction of decision trees, it presents the novelty of receiving structures (subgraphs) immersed in a property graph as training set. By using appropriate predicates, GGQs constructed by the algorithm can not only evaluate properties of the structure under evaluation, but also properties of any element (subgraph) in its environment.

The complete training set, $ \mathcal {L} $, will consist of pairs of the form $ (S, value) $, where $ S $ is a subgraph of $ G $ and $value$ is its associated output value. Consequently, each node $n$ of the decision tree will have associated the following structures:

\begin{itemize}
\item $\mathcal{L}_n = \{(S_1,y_1),...,(S_N,y_N)\} \subseteq \mathcal{L}$, a subset of the training set.
\item $Q_n = (V_{Q_n},E_{Q_n},\alpha_{Q_n},\theta_{Q_n})$, A GGQ verified by all subgraphs in $ \mathcal {L}_n $. 
\end{itemize}

Algorithm \ref{alg:graph-id3} formally represents GGQ-ID3. Note that the set of available refinements, $ REFS $, to extend the GGQ in each step, as well as the stopping condition and refinement selection criteria, remain as free parameters of the model.

\begin{algorithm}
\begin{algorithmic}[1]
\STATE {Create a $Root$ node for the tree}
\IF {Stop criteria is reached}
\RETURN The single-node tree $Root$, with most frequent label in $\mathcal{L}$.\label{op0}
\ELSE
\STATE {$(Q_1,..,Q_k) = Optimal\_Refinement(G,Q,L,REFS)$}
\STATE {$\mathcal{L}_1= \{(S,y) \in \mathcal{L} : S\vDash Q_1\},...,\mathcal{L}_k= \{(S,y) \in \mathcal{L} : S\vDash Q_k\}$}
\STATE {Add $k$ new tree branches below $Root$ with values GGQ-ID3($G,Q_i,\mathcal{L}_i,REFS$) for every $1\leq i \leq k$.}
\ENDIF
\end{algorithmic}
\caption{GGQ-ID3($G,Q,\mathcal{L},REFS$)}\label{alg:graph-id3}
\end{algorithm}

In a typical learning process, the input of the algorithm will be:

\begin{itemize}

\item $G = (V,E,\mu)$, graph in which the structures to be classified are immersed.

\item $\mathcal{L}= \{(S_1,y_1),...,(S_N,y_N)\}$, set of pairs $ (S_i, y_i) $ where $ S_i \subseteq G $ represents a subgraph and $ y_i $ is its associated output value. 

\item $Q_0 = (V_Q,E_Q, \mu_Q)$, initial GGQ (usually a GGQ with the largest common structure in $ S_1, ..., S_N $).

\item $REFS$, available refinements set.

\end{itemize}

The algorithm follows the usual procedure in ID3. It begins by creating a tree containing a single node (root node) to which all the $ \mathcal {L} $ objects and $ Q_0 $ are associated. If $ n $ is the node of the decision tree which we are working with, the algorithm evaluates which refinement divides $ \mathcal{L}_n $ better and it will be chosen to be applied to $ Q_n $. Next, as many branches from $ n $ as GGQs are in the chosen refinement will be created and each pair of $ \mathcal{L}_n $ will be transmitted through the corresponding branch. Each child node of $ n $ will inherit the GGQ resulting from the refinement and proceed to search (if the stop condition is not reached) the best refinement for the new GGQ. Each branch of the tree will receive a set of objects and a GGQ that is verified by all of them. If the stop condition is met, the node will become a leaf node associated with the corresponding class. This process is repeated recursively.

Decision trees obtained in most automatic procedures divide the data into binary complementary subsets in a recursive way \cite{c45}, but in the multi-relational case the production of complementary patterns of this type is not straightforward. This is the reason a set of refinements that generate complementary GGQs, but not necessarily binary, were presented. 

\section{Some Examples}

Next we present an example of GGQ-ID3 algorithm application on a small property graph as a demonstration. The available refinements $REFS$ will be those seen in the previous section, we will use the most restrictive stop condition (all the pairs in the current node belong to the same class), and the Information Gain \cite{Mitchell:1997:ML:541177} will be used as impurity measure.

\begin{figure}[h!]
  \centering
  \includegraphics[width=\textwidth]{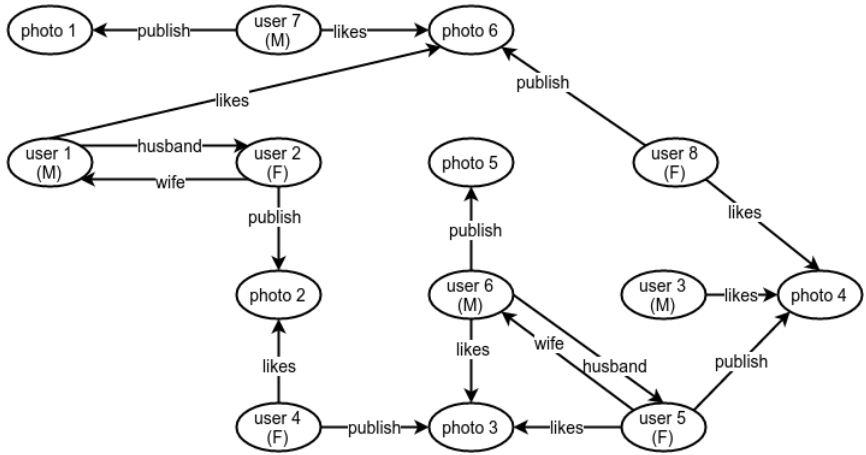}
  \caption{Social Graph}
  \label{grafo1}
\end{figure}

We will work with the social graph shown in Figure \ref{grafo1}, which represents some marital connections between users and information related to photographs publication. There are nodes of types \texttt{user} and \texttt {photo}, and edges of types \texttt{wife}, \texttt{husband}, \texttt{publish} and \texttt{likes}. In this case, $ L $ is not extended with topological measures of the graph. In addition, \texttt{user} nodes have gender property with value \texttt{F} (female) or \texttt{M} (male). \texttt{photo} nodes have the value \texttt{None} associated to gender property. 

Although GGQ-ID3 algorithm is able to construct decision trees to classify any structure (subgraph), in order to show a simple example, we approach a problem of classifying nodes, specifically trying to predict the gender of nodes. In order to make the exposition clearer and to avoid confusion, the nodes belonging to the training set will be called \emph{objects}, and we will use \emph{nodes} for the nodes of the decision tree under construction.

\begin{figure}[h!]
	\centering
	\includegraphics[scale=0.25]{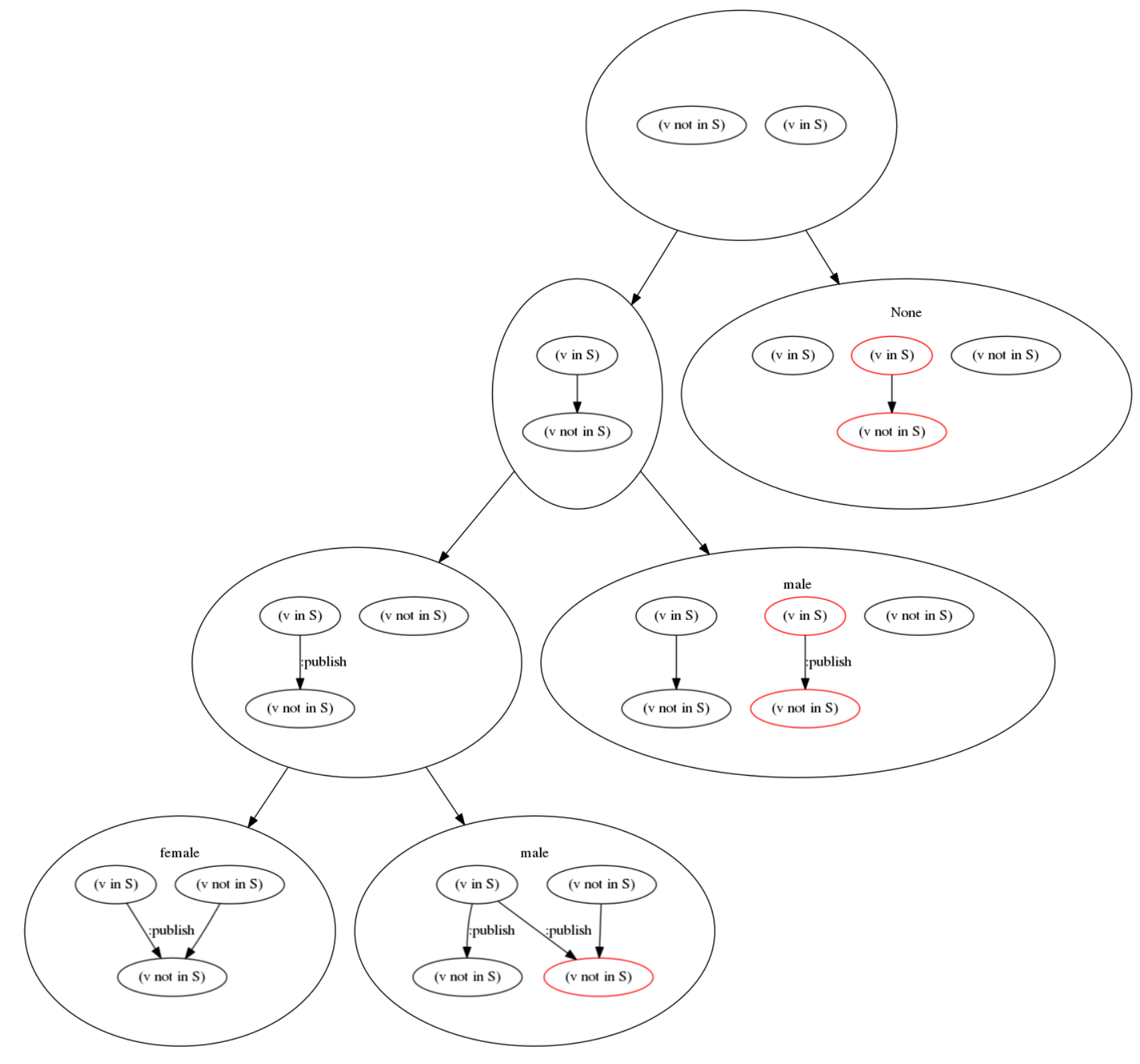}
	\caption{GGQ Decision Tree (social graph)}
	\label{social1}
\end{figure}

GGQ-ID3 algorithm constructs the decision tree in Figure \ref{social1} (positive nodes/ edges are marked in black and negative ones in red) that correctly classify all nodes from Figure \ref{grafo1} according to their gender: \texttt{M}, \texttt{F} or \texttt{None}. Next indications about the  construction follow the representation shown in Figure \ref{social1}.

The initial training set, $ \mathcal{L} $, is formed by all pairs of nodes and their corresponding gender. Initial GGQ, $ Q_0 $, is composed of two positive nodes, one with a predicate that requires belonging to the subgraph under evaluation ($ v \in S $) and another with the opposite condition ($ V \notin S $). As previously mentioned, usually the initial GGQ will correspond to the largest common substructure in the subgraphs to be classified, in this case the largest common structure is composed of a single node with no restrictions. In addition, and for reasons of efficiency, in each step of the algorithm an isolated positive node will be created with a predicate that requires non belonging to the subgraph under evaluation ($ v \notin S $) if there is no such isolated node. If the subgraph under evaluation does not cover all the nodes of the graph in which it is immersed (which is true in all the examples presented), adding a node of this type does not modify the meaning of the GGQ but allows to reach optimal GGQ easily.
 
As a first step, the algorithm will analyse what refinement in $REFS$ (and with what parameters) represents the highest impurity reduction, resulting in refinement $+\{\overset {e^+}{\longrightarrow}\}$, add edge between the only two existing nodes in $Q_0$. Although refinement \textbf{add edge} generates a refinement of four elements, we consider only those that intervene in the classification process, removing those that transmit an empty set of objects.

Because the subgraphs to classify are composed of a single object, this refinement evaluates the existence or not of an outgoing edge in this object. According to Figure \ref{social1}, the right branch will transmit all objects with no outgoing relations. In this case, all \texttt{photo} objects of the data graph, that have \texttt{None} as \texttt{gender}, so this branch is associated with the output value \texttt {None} and becomes a leaf of the decision tree. 

The left branch will transmit all objects with an outgoing relation (in this case, all objects of the data graph of type \texttt{user}). As the values of \texttt{gender} are not homogeneous in this set, the algorithm must continue and a new refinement to this branch have to be applied. Here we have nodes that reflect male and female users with an outgoing edge.

$ +\{ \overset {e \wedge \{type = publish \}} {\longrightarrow} \} $ refinement (\textbf{add a new predicate to edge}) produces the greatest information gain. The refinements applied to this node in the decision tree will discriminate which objects (of those with an outgoing edge) verify that this edge is of type \texttt{publish} and which do not. According to Figure \ref{social1}, objects with no outgoing edge of type \texttt{publish} will be transmitted through the right branch of this node. In our case, all these objects are male gender users, so the node becomes a leaf associated with class \texttt{M}. Through the left branch the objects with an outgoing \texttt{publish} edge will be transmitted. In this case, again the values of the \texttt{gender} property are not homogeneous (they present impurity) so the algorithm must continue looking for a new refinement in this branch.

We repeat the process for the new node, with objects having an outgoing edge of type \texttt{publish}. The refinement with the maximum impurity reduction is $ + \{ \overset {e^+} { \longrightarrow} \} $, \textbf{add an edge} between the isolated node ($ v \notin S $) and the target node of the \texttt{publish} relationship. The interpretation is to discriminate, from the objects that have published something, those whose publications receive some incoming edge by another object outside the structure under evaluation. In other words, users that have published a photo that someone else likes.

All objects in the right branch are associated to gender \texttt{M} (users who have posted a photo that nobody likes), so the stop condition is reached and the node becomes a leaf associated with class \texttt{M}. Users who have published a photo and there is someone who likes it will be transmitted by the left branch. All these users are of gender \texttt{F}, so the stop condition is also checked and the produced leaf is associated with the corresponding class.

We have obtained a decision tree able to correctly classify all the objects in the data graph regarding to their gender making use of the multi-relational structure in which they are immersed.
In addition, we automatically obtain GGQ that evaluate properties of the context differentiating between the objects that must be inside the structure under analysis and those that must be outside, one of the big differences that GGQ shows in relation with other multi-relational query frameworks. 

Let us continue showing some more examples of multi-relational decision trees that use GGQ and that have been obtained following the GGQ-ID3 algorithm. As in the case explained above, isolated positive nodes non belonging to the subgraph under evaluation were added at each step if they did not exist in the GGQ. These examples have been extracted from small databases with sufficient complexity to show pattern discovery capability of the GGQ-ID3 algorithm. In some cases only the more interesting leafs are shown.

\subsection{StarWars}

\begin{figure}[htb]
    \begin{center}
        \includegraphics[width=\textwidth]{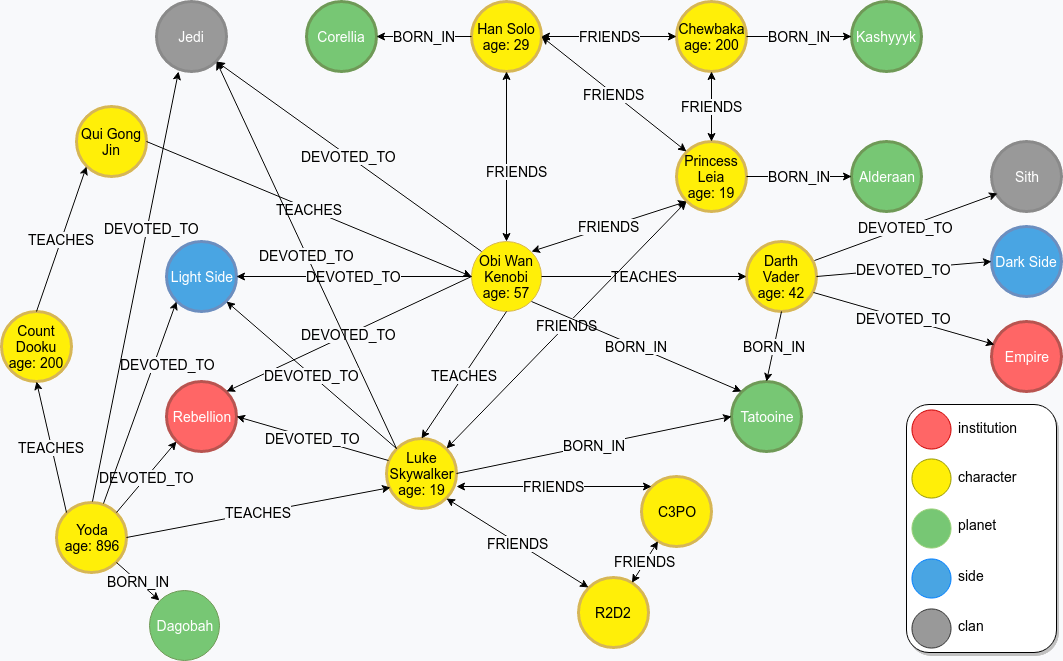}
    \end{center}
    \caption{%
        Starwars Graph
    }%
    \label{starwars}
\end{figure}

In this case we will mine the graph presented in the Figure \ref{starwars} with information on StarWars \footnote{\url{http://console.neo4j.org/?id=StarWars}}.

\begin{figure}[htb]
  \centering
  \includegraphics[width=\textwidth]{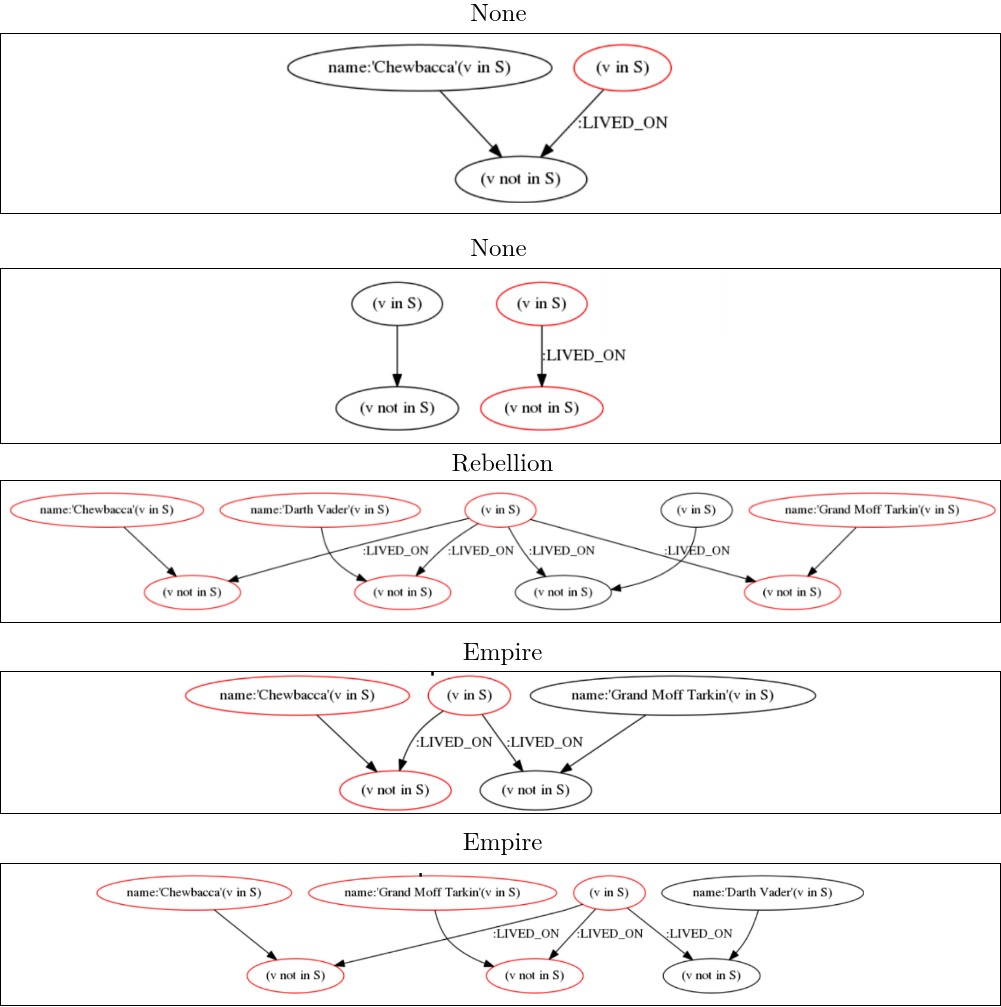}
  \caption{GGQ Decision Tree Leafs (Starwars graph)}
  \label{starwars1}
\end{figure}

The several GGQ shown in Figure \ref{starwars1} allow to discriminate each character in the graph according to whether it is devotee of the Empire, Rebellion or neither. They correspond to classification leafs of the decision tree automatically generated by GGQ-ID3. To construct it, nodes of type institution (along with the edges in which they participate) have been removed from the original graph database. The queries show that even working with small graphs a high semantic query level is automatically obtained.

\subsection{The Hobbit}

\begin{figure}[htb]
  \centering
  \includegraphics[width=\textwidth]{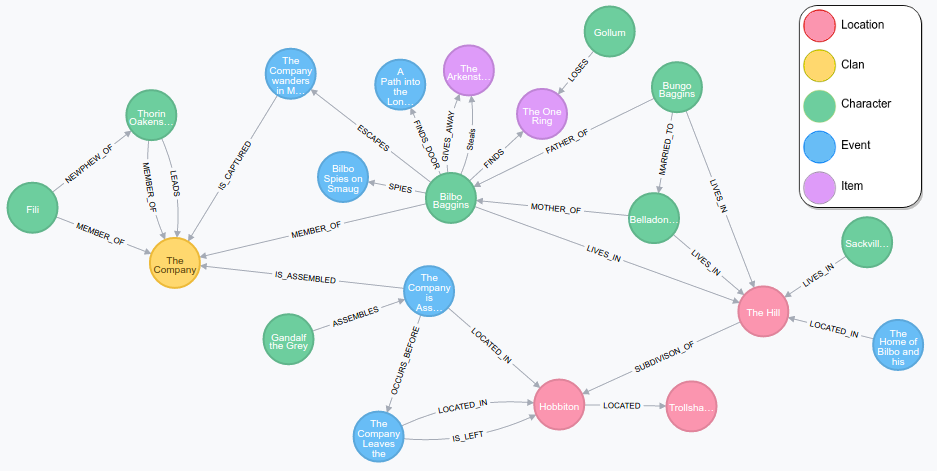}
  \caption{Hobbit Graph (section)}
  \label{ejhobbit}
\end{figure}

The third example is obtained by mining another toy graph, in this case related to the the Lord of the Rings world \footnote{\url{http://neo4j.com/graphgist/c43ade7d259a77fe49a8}}. This graph contains 105 nodes with 7 different types (\texttt{Character}, \texttt{Event}, \texttt{Item}, \texttt{Clan}, \texttt{Aligment}, \texttt{Location} and \texttt{Text}) and 209 edges distributed through 65 types, showing a very high edge typology, with very few instances of some types of edges, so that inefficient mechanisms will tend to create very large trees. Figure \ref{ejhobbit} shows a subgraph extracted from this database.

\begin{figure}[htb]
  \centering
  \includegraphics[width=\textwidth]{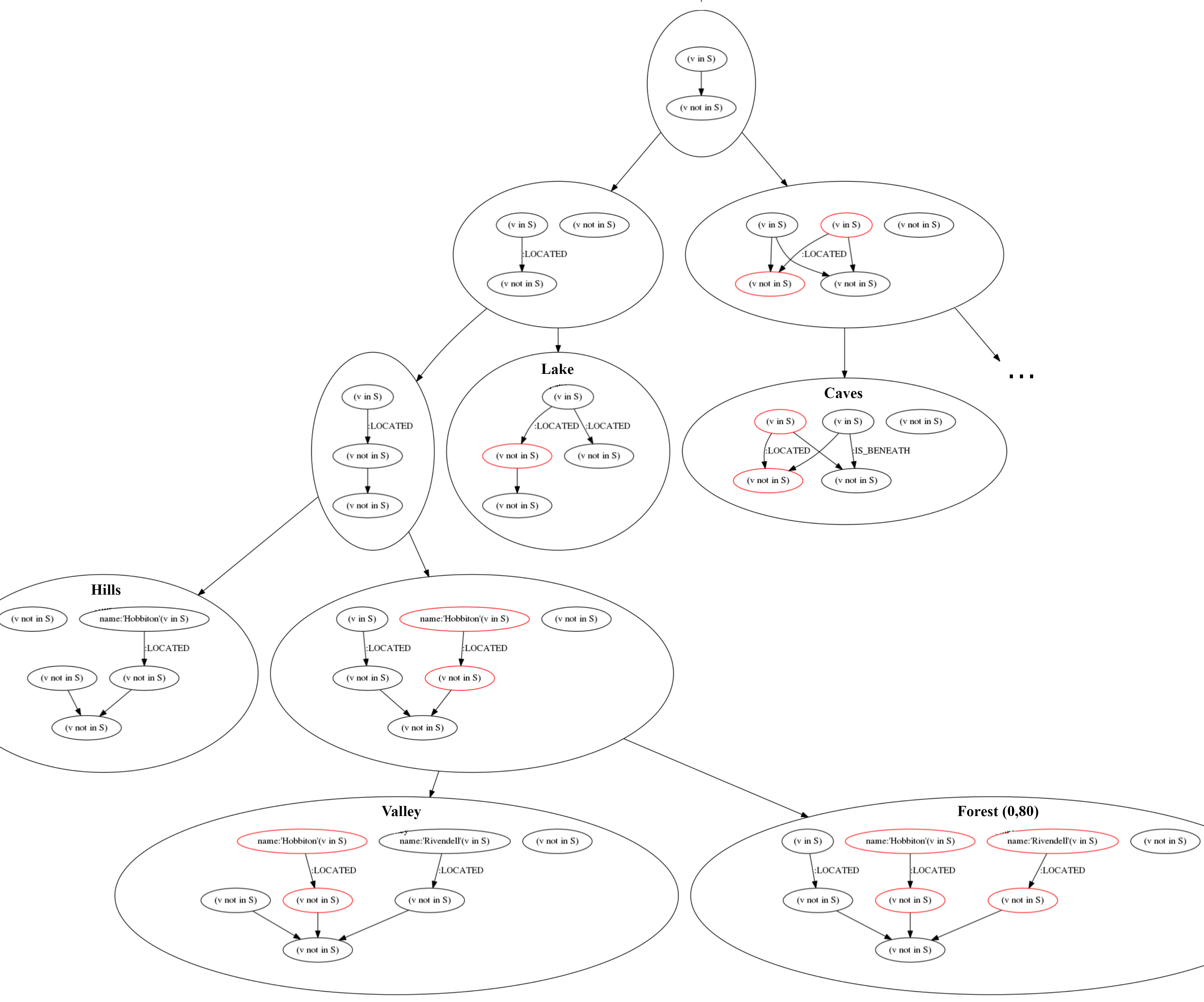}
  \caption{GGQ Decision Tree (Hobbit graph)}
  \label{hobbit1}
\end{figure}

Decision tree presented in Figure \ref{hobbit1} was automatically obtained using GGQ-ID3 algorithm and allows to discriminate location types (\texttt{Hills}, \texttt{Fo\-rest}, \texttt{Valley}, \texttt{Mountain}, \texttt{Caves} and \texttt{Lake}). A maximum depth of 5 levels was imposed in the construction of the tree.

\section{Some Notes About Implementation}

In order to perform verification tests on the GGQ query system and GGQ-ID3 algorithm, a proof-of-concept implementation \footnote{\url{https://github.com/palmagro/ggq}} developed in Python and using Neo4j \footnote{\url{https://neo4j.com/}} as persistence system has been carried out. Some parts of the system has been implemented using the Cypher query language. This implementation would gain efficiency if, instead of using Cypher, the Java API of Neo4j or an \textit{ad hoc} persistence system oriented to support this type of tasks would been used. 

Although our goal has been to demonstrate that the formal query system is able to perform this type of tasks in a simple way, some basic optimizations have been made in the implementation (like the use of complementarity of the queries from a refinement to save time and empty leaves pruning).

If an object verifies a GGQ, it will also verify all its predecessors GGQ, so the classification power of a GGQ-ID3 decision tree is contained in the leaves. To gain efficiency, when classifying a new example, is advisable to use the complete tree, since only at least $k$ evaluations (with $k$ the depth of the tree) are needed, but there may be an exponential amount of classification leaves. In addition, since an atomic change has occurred from one level to the next, provided by the corresponding refinement, the evaluation of a subgraph in a node of the decision tree involves considering only some additional checks added to the evaluation of its parent node.

\section{Conclusions and Future Work}

In this paper the GGQ-ID3 algorithm has been presented, it uses Generalized Graph Query framework as test tools following the fundamentals of the algorithm ID3. GGQ-ID3 has shown capabilities to extract interesting patterns to be used in complex learning tasks, that can be interpreted as new attributes discovered by the algorithm (\textit{feature extraction}).

We have used an initial set of refinements in the examples, but this set can be modified by adding refinements according to the structure of the graph under evaluation or to the task to be achieved.

MRDTL algorithm was developed a decade ago to work specifically on relational databases and with simple classification tasks, and can be seen as a particular case of GGQ-ID3 algorithm where only tree-shaped GGQ are allowed (since they use selection graphs) and where only learning from very simple structures (nodes) is allowed. In this sense, GGQ-ID3 is a step forward in this line of work.

The main problem presented by multi-relational decision tree construction algorithms is that the hypothesis space is extremely large. GGQ-ID3 also suffers from this problem. To reduce complexity, and to guide the search, several solutions can be proposed. On the one hand, the frequency of certain structures can be statistically analysed in order to reduce the number of possible refinements to be applied in each case and the refinements search cost. Thus, it is necessary to use measures developed for generalized graphs extending the simpler frequency measurements that are used in the case of MRDTL-2 \cite{Atramentov2003}. On the other hand, more complex refinement families (for example, combining \textbf{add edge} with \textbf{adding property to an edge} in a single step) can be created to reduce the number of steps to get complex GGQ. If this last option is carried out properly (unifying the refinements according to the frequency of structures in the graph), the algorithm can reach the solution faster. In both cases, an improvement in efficiency is achieved by sacrificing the possibility of covering a wider hypothesis space (but probably offering alternatives in which the impurity reduction is larger). In this sense, a minimal set of well-constructed refinements has been offered in this paper, but not with the intention of being optimal for certain learning tasks.

The second major problem with GGQ-ID3 (shared by all ID3-inspired algorithms) is the inability to undo the decisions taken during the construction process. Refinement election at a given step depends on refinements chosen in previous steps. To solve this problem, some backtracking procedure can be considered, or some Beam-Search procedure (as in the algorithm GBI \cite{Geamsakul}), allowing to take several decisions in parallel, and finally select the one that has resulted in a better solution.

With regard to the future work that derives from the research presented here, it is worth mentioning that, since GGQs are constructed using the generalized graph structure, and that this structure allows the definition of hypergraphs in a natural way, GGQs can evaluate hypergraphs with properties just with slight modifications. In addition, the development of different refinement sets according to the type of graph to query or even the automatic generation of such sets from statistics extracted from the graph to be analysed can lead to important optimizations in processes of automatic and effective GGQ construction.

Decision trees are an ideal tool to be combined through some ensemble model, such as Random Forest, thanks to its low computational training cost and the randomness obtained from small changes in the data set from which to learn. Therefore, having adequate models for automatic generation of multi-relational random forests becomes a task of great importance.

Mulri-relational Machine Learning has received (and still receives) less attention than the more standard machine learning methods that make use of non relational information (usually in the form of tables and other more regular structures). The most commonly used databases, in which information about most of the studied phenomena is stored, make use of schemes and systems based on the relational model that do not show optimal performance when working with complex relationships. In addition, the greater expressive richness of more complex information representation structures imposes greater difficulty in making new algorithms and provides, at least in the first approximations, less striking results than the more refined and more traditional methods.

Another important feature of decision tree-based learning methods is that they represent a white-box model, providing an interpretable explanation of the decisions taken when performing regression or classification. In the case of GGQ-ID3, this characteristic is enhanced due to the interpretability of the GGQ. This advantage may be blurred when using ensembles, but there are methods to extract knowledge of aggregated results that could be reused in this context. One possibility is to combine the leaves associated with the same class from GGQ of different trees, giving rise to combined patterns (possibly probabilistically) that are able to condense the different predicates that characterize the same class in a more powerful predicate.

\bibliographystyle{plain}	
\bibliography{biblio}

\end{document}


\title{Inducción de Árboles de Decisión basados en Generalized Graph Queries}
\author{Pedro Almagro-Blanco and Fernando Sancho-Caparrini}

\maketitle

\section{Introducción}

Un árbol de decisión es un modelo de clasificación (y regresión) que, a partir de las características de un objeto dado, y aplicando una serie de reglas, es capaz de asociar una clase a dicho objeto (o un valor continuo en el caso de regresión). La inducción de árboles de decisión a partir de un conjunto de objetos previamente clasificados es uno de los modelos de aprendizaje automático más populares debido, entre otras cosas, a la baja demanda computacional en su entrenamiento y a la facilidad de interpretación de sus resultados por parte de los humanos, por lo que se coloca como representante de los llamados modelos de caja blanca.

El algoritmo ID3 fue ideado por R. Quinlan en 1983 \cite{id3} para crear árboles de decisión clasificadores a partir de un conjunto de datos conformado por una serie de \textit{objetos} descritos a través de una colección de \textit{propiedades}. Cada objeto del conjunto de entrenamiento pertenece a una \textit{clase} (normalmente, representado por medio del valor del atributo \textit{target}) de un conjunto de clases mutuamente excluyentes. ID3 ha sido una de las técnicas más utilizadas de aprendizaje automático, con aplicaciones a tareas tan diversas como predicción de epidemias, control de robots o clasificación automática de clientes en bancos o entidades aseguradoras \cite{takale2004constructing,Salganicoff1996,He2012}.

El objetivo de este trabajo es ofrecer una metodología que permita llevar a cabo tareas de aprendizaje automático haciendo uso de árboles de decisión sobre datos en forma de grafo. En este contexto, el número de posibles propiedades de cada objeto va mucho más allá de aquellas que tiene asociadas directamente, ya que las propiedades de los elementos que se relacionan con él también pueden ser considerados atributos del mismo, es más, incluso la estructura topológica que forman los objetos de su entorno y las diversas medidas que se pueden tomar de esta estructura podrían ser consideradas como atributos. Con este objetivo, analizaremos diferentes técnicas que permiten la construcción automática de árboles de decisión a partir de datos estructurados en forma de grafo y presentaremos nuestra propuesta, GGQ-ID3, que tiene como objetivo proporcionar un marco para clasificar subestructuras en un grafo, desde simples nodos y aristas, hasta caminos y subgrafos de mayor envergadura, haciendo uso del concepto de Generalized Graph Query (GGQ, para abreviar, a partir de ahora) \cite{2017arXiv170803734A}.

Este artículo se estructura de la siguiente manera: realizaremos un repaso por las diferentes técnicas existentes de inducción de árboles de decisión relacionales (aquellos que trabajan sobre estructuras de datos en las que hay relaciones entre sus elementos) y a continuación daremos una serie de definiciones previas que nos permitirán presentar nuestra propuesta basada en el uso de Generalized Graph Queries como herramienta de evaluación en los nodos del árbol. Una vez planteada nuestra propuesta, mostraremos algunos ejemplos de aplicación de la misma. Cerraremos el trabajo presentando algunas conclusiones y líneas futuras que han ido surgiendo durante la ejecución del mismo.

\section{Trabajos Relacionados}
\label{mrdtl}
 
Este apartado no pretende ser una recopilación exhaustiva de los trabajos relacionados que se pueden encontrar en la literatura, sino una selección de aquellos que hemos encontrado más interesantes, bien sea por su capacidad de predicción, su eficiencia computacional, o porque han servido de base a otras técnicas y trabajos relevantes.

La \emph{Programación Lógica Inductiva} (ILP) \cite{plotkin1972automatic} es un área del aprendizaje automático que utiliza fundamentos de Programación Lógica para representar de manera uniforme ejemplos, base de conocimientos, e hipótesis. A pesar de que la ILP por sí misma (sin una transformación adecuada de las relaciones entre datos a predicados lógicos) no permite generar árboles de decisión relacionales, sí permite generar de manera automática árboles de decisión lógicos que pueden ser considerados la base de uno de los algoritmos más importantes de generación automática de árboles de decisión, como es el algoritmo Multi-Relational Decision Tree Learning (MRDTL). La gran potencia que proporciona ILP en nuestro contexto es su interpretabilidad, pero su punto débil radica en la ineficiencia para trabajar con bases de datos complejas \cite{DBLP:journals/corr/abs-1211-3871}.

Un \emph{Árbol de Decisión Lógico} \cite{ltrees} es un árbol de decisión binario en el que que todos los test de los nodos internos se expresan como conjunción de literales de una Lógica de Primer Orden prefijada. El algoritmo TILDE (Top-Down Induction of Logical Decision Trees) \cite{ltrees} construye árboles de decisión lógicos para clasificar instancias a partir de un conjunto de ejemplos clasificados, una base de conocimientos, y un lenguaje que indica qué tipo de preguntas están permitidas en el árbol \cite{BLOCKEEL1998285}. La única diferencia entre este algoritmo y el ID3 presentado por Quinlan (sin tener en cuenta las posibles optimizaciones implementadas en C4.5 u otros) se encuentra en los tests llevados a cabo en cada nodo del árbol. Tras el auge de la ILP se lograron algunos avances importantes en la minería de datos multi-relacional \cite{Kavurucu_confidence-basedconcept,10.1109/GCIS.2009.289}. Yin Xiaoxin \cite{Yin2006} diseñó \emph{CrossMine}, un modelo de clasificación multi-relacional que mezcla ILP y las bases de datos relacionales, mejorando la eficiencia en este tipo de tareas a través de un método para realizar uniones virtuales de tablas de la base de datos \cite{improved}.

\emph{Multi-Relational Decision Tree Learning} (MRDTL) es un algoritmo para el aprendizaje de árboles de decisión multi-rela\-cio\-na\-les \cite{Leiva02mrdtl:a} basado en las ideas de Knobles et al. \cite{Knobbe99multi-relationaldecision} que trabaja con el concepto de Grafo de Selección (Selection Graph). Un grafo de selección es una representación en forma de grafo de una consulta SQL que selecciona registros que cumplan con una serie de restricciones expresadas en forma de estructuras con las que debe (o no debe) estar conectado el registro bajo evaluación. Además, los grafos de selección permiten ser refinados usando un conjunto de operaciones atómicas con el fin de construir grafos de selección complejos a partir de un grafo de selección inicial. Además, para poder ser usados en un proceso similar a ID3, se debe poder obtener el grafo de selección complementario a uno dado. Esencialmente, MRDTL funciona como ID3, pero se caracteriza por hacer uso de grafos de selección como atributos binarios en cada nodo de decisión del árbol. La especificación de este método está orientada a bases de datos relacionales debido, en parte, a que en el tiempo en el que se presentó aún no se habían desarrollado otro tipo de bases de datos más adecuadas para este tipo de tareas. 

\emph{Graph-Based Induction} (GBI) es una técnica de minería de datos que extrae patrones frecuentes (\textit{network motifs}) de grafos etiquetados y dirigidos a través de la unión de pares de nodos conectados \cite{Nguyen2005} y que es muy eficiente debido a que utiliza una técnica voraz. A partir de esta técnica, \emph{Decisión Tree Graph-Based Induction} (DT-GBI) es un algoritmo de construcción de árboles de decisión para clasificar grafos utilizando los principios de GBI. En DT-GBI los atributos (llamados patrones o subestructuras) son generados durante la ejecución del algoritmo \cite{Geamsakul2003}, por lo que DT-GBI es un generador de árboles de decisión con capacidad de construcción de atributos \cite{Nguyen2006}. Hay que indicar que, a diferencia de nuestra propuesta o de MRDTL, DT-GBI construye árboles de decisión para clasificar grafos completos, y no subestructuras generales inmersas en él (como es el caso de GGQ-ID3) o nodos (como es el caso de MRDTL).

A continuación presentamos brevemente el concepto de Generalized Graph Query, base del modelo de generación de árboles de decisión GGQ-ID3, se puede encontrar una descripción más amplia del mismo en \cite{2017arXiv170803734A}. 

\section{Generalized Graph Query}
\label{PQG}

El \emph{Grafo Generalizado} (que a veces, y por extensión, también denominaremos \emph{Grafo con Propiedades}) es un concepto que abarca diferentes variantes de grafos que se pueden encontrar en la literatura, tanto aquellos que sirven desde un punto de vista puramente matemático, como los que sirven de sustrato teórico para las redes semánticas o bases de datos en grafo.

\begin{definition}
Un \emph{Grafo Generalizado} es una tupla $G = (V,E,\mu)$ donde:
\begin{itemize}
	\item $V$ y $E$ son conjuntos, que llamaremos, respectivamente, \emph{conjunto de nodos} y \emph{conjunto de aristas} de $G$.
	\item $\mu$ es una función que asocia a cada nodo o arista en el grafo su conjunto de propiedades, es decir, $\mu : ( V \cup E) \times R \rightarrow S$, donde $R$ representa el conjunto de posibles \emph{claves} para dichas propiedades, y $S$ el conjunto de posibles \emph{valores} asociados a las mismas.
\end{itemize} 

Habitualmente, para cada $\alpha \in R$ y $x\in V\cup E$, escribiremos $\alpha(x)=\mu(x,\alpha)$.

Además, exigiremos la existencia de una clave destacada para las aristas del grafo, que llamaremos \emph{incidencias} y denotaremos por $\gamma$, que asocia a cada arista del grafo una tupla, ordenada o no, de vértices del grafo.
\end{definition}

Las \emph{Generalized Graph Queries} permiten llevar a cabo consultas estructurales y semánticas, exactas, óptimas, y basadas en un tipo de Regular Pattern Matching que permite, además de proyectar aristas del patrón en caminos (no necesariamente aristas) que cumplan las restricciones impuestas, expresar restricciones más complejas sobre cada elemento del patrón y realizar consultas que posean ciclos.

Si consideramos $L$ un Lenguaje de Primer Orden con igualdad que usa como símbolos no lógicos una colección que contiene todas las funciones de $\mu$ junto con constantes asociadas a cada elemento del grafo (y algunos símbolos adicionales, por ejemplo para denotar métricas definidas sobre los elementos del grafo), y construimos de la forma usual el conjunto de términos del lenguaje y el conjunto de fórmulas, $FORM(L)$ (que llamaremos predicados), podemos definir las consultas sobre grafos generalizados haciendo uso de las mismas estructuras como:

\begin{definition}
    Un \emph{Generalized Graph Query (GGQ)} sobre $L$ es un grafo generalizado, $Q = (V_Q,E_Q,\mu_Q)$, donde existen $\alpha$ y $\theta$, propiedades destacadas en $\mu_Q$, tales que:
    \begin{itemize}
        \item $\alpha:V_Q\cup E_Q\rightarrow \{+,-\}$ total.
        \item $\theta:V_Q\cup E_Q\rightarrow FORM(L)$ asocia un predicado binario, $\theta_x$, a cada elemento $x$ de $V_Q\cup E_Q$.
    \end{itemize}
\end{definition}

Como veremos, la segunda entrada de estos predicados binarios se usará para hablar de condiciones de pertenencia sobre subgrafos de $G$ (el grafo general sobre el que estamos evaluando las consultas), mientras que la primera esperará recibir como entrada elementos adecuados al tipo de elemento al que está asociado. 

Dado un GGQ en las condiciones anteriores, notaremos $x^+$, respectivamente $x^-$, para indicar que $\alpha(x)=+$, respectivamente $\alpha(x)=-$. Si para un elemento $x$, $\theta_x$ no está explícitamente definida, supondremos que $\theta_x$ es una tautología, que podemos denotar en general por $T$. Intuitivamente los elementos positivos del patrón representan elementos que deben estar presentes en el grafo sobre el que se realiza la consulta y que verifican los predicados asociados, mientras que los elementos negativos en el patrón representan elementos que no pueden estar presentes en el grafo.

Para poder expresar con más facilidad las condiciones necesarias que definen la aplicación de un GGQ sobre un grafo, así como los resultados que veremos más adelante, introducimos las notaciones:

\begin{definition}
    Dado $Q=(V_Q,E_Q,\mu_Q)$ un GGQ, el conjunto de $Q$-predicados asociados a $Q$ es:
    \begin{enumerate}
        \item Para cada arista, $e\in E_Q$:
        $$Q_{e^o}(v,S)=\exists \rho\in \mathcal{P}_v^o(G)\ \left(\theta_e(\rho,S)\wedge \theta_{e^o}(\rho^o,S) \wedge \theta_{e^i}(\rho^i,S)\right)$$
        $$Q_{e^i}(v,S)=\exists \rho\in \mathcal{P}_v^i(G)\ \left(\theta_e(\rho,S)\wedge \theta_{e^o}(\rho^o,S) \wedge \theta_{e^i}(\rho^i,S)\right)$$

        En general, escribiremos $Q_{e^*}(v,S)$, donde $*\in \{o,i\}$, y notaremos:
        $$Q_{e^*}^+ = Q_{e^*},\hspace{1cm} Q_{e^*}^- = \neg Q_{e^*}$$
        \item Para cada nodo, $n\in V_Q$:
        $$Q_n(S)=\exists v\in V\ \left(\bigwedge_{e\in \gamma^o(n)} Q_{e^o}^{\alpha(e)}(v,S)\ \wedge \bigwedge_{e\in \gamma^i(n)} Q_{e^i}^{\alpha(e)}(v,S)\right)$$
        Además, notaremos:
        $$Q_n^+ = Q_n,\hspace{1cm} Q_n^- = \neg Q_n$$
    \end{enumerate}
\end{definition}

Donde $e^o$ representa el nodo del que parte la arista $e$ y $e^i$ representa el nodo destino de dicha relación, $\mathcal{P}_v^o(G)$ (resp. $\mathcal{P}_v^i(G)$) representa los caminos en $G$ que parten del (resp., terminan en) nodo $v$.

A partir de estas notaciones, podemos definir formalmente cuándo un subgrafo verifica un GGQ determinado:

\begin{definition}
\label{verifica}
    Dado un subgrafo $S$ de un grafo con propiedades, $G=(V,E,\mu)$, y un Generalized Graph Query, $Q=(V_Q,E_Q,\mu_Q)$, ambos sobre el lenguaje $L$, diremos que $S$ \emph{verifica} $Q$, y lo denotaremos $S\vDash Q$, si se verifica la fórmula:

    $$Q(S)=\bigwedge_{n\in V_Q} Q_n^{\alpha(n)}(S)$$

    En caso contrario, escribiremos: $S\nvDash Q$.
\end{definition}

Con el fin de obtener métodos controlados de generación de consultas se pueden definir refinamientos para ir modificando un GGQ por pasos unitarios. Dados dos GGQ, $Q_1,\ Q_2$, $Q_1$ \emph{refina} $Q_2$ en $G$, $Q_1\preceq_G Q_2$, si para todo $S\subseteq G$, si $S\vDash Q_1$, entonces $S\vDash Q_2$.

\begin{definition}
    Dado $Q\in GGQ$. Diremos que un conjunto de GGQs $R$ es un \emph{conjunto de refinamiento} de $Q$ en $G$ si verifica:
    \begin{enumerate}
        \item $\forall\ Q'\in R\ (Q'\preceq_G Q)$
        \item $\forall\ S\subseteq G\ (S\vDash Q\Rightarrow \exists !\ Q'\in R\ (S\vDash Q'))$
    \end{enumerate}
\end{definition}

En lo que sigue, dado $Q$, $Cl_Q^{W}$ representa un grafo derivado de $Q$ en el que se han duplicado los nodos presentes en $W \subseteq V_Q$ (con las respectivas aristas si las tuvieran). 

Siguiendo \cite{2017arXiv170803734A}, se puede probar que los siguientes conjuntos de GGQ son refinamientos de $Q$:
\begin{itemize}
	\item \textbf{Añadir nodo nuevo}: si $m\notin V_Q$, entonces $Q+\{m\}$:
        \begin{align*}
        Q_1 &= (V_Q\cup\{m\},\ E_Q,\ \alpha_Q\cup(m,+),\ \theta_Q\cup(m,T))\\
        Q_2 &= (V_Q\cup\{m\},\ E_Q,\ \alpha_Q\cup(m,-),\ \theta_Q\cup(m,T))
        \end{align*}
	\item \textbf{Añadir arista nueva entre nodos positivos}: si $n,m\in V_Q^+$, entonces $Q+\{n^+\overset {e^*}{\longrightarrow} m^+\}$ ($*\in\{+,-\}$) (donde $Q'=Cl_Q^{\{n,m\}}$):
		\begin{align*}
		Q_1 &= (V_{Q'},\ E_{Q'}\cup\{n^+\overset {e^*}{\longrightarrow} m^+\},\ \theta_{Q'}\cup(e,T))\\
		Q_2 &= (V_{Q'},\ E_{Q'}\cup\{n^+\overset {e^*}{\longrightarrow} m^-\},\ \theta_{Q'}\cup(e,T))\\
		Q_3 &= (V_{Q'},\ E_{Q'}\cup\{n^-\overset {e^*}{\longrightarrow} m^+\},\ \theta_{Q'}\cup(e,T))\\
		Q_4 &= (V_{Q'},\ E_{Q'}\cup\{n^-\overset {e^*}{\longrightarrow} m^-\},\ \theta_{Q'}\cup(e,T))
		\end{align*}
	\item \textbf{Añadir predicado a arista positiva entre nodos positivos}: si $n,m\in V_Q^+$, con $n^+\overset {e^+}{\longrightarrow} m^+$, y $\varphi\in FORM$, entonces $Q+\{n^+\overset {e \wedge \varphi}{\longrightarrow} m^+\}$ (donde $Q'=Cl_Q^{\{n,m\}}$):
		\begin{align*}
		Q1 &= (V_{Q'},\ E_{Q'}\cup\{n^+\overset {e'}{\longrightarrow} m^+\},\ \theta_{Q'}\cup(e',\theta_e\wedge \varphi))\\
		Q2 &= (V_{Q'},\ E_{Q'}\cup\{n^+\overset {e'}{\longrightarrow} m^-\},\ \theta_{Q'}\cup(e',\theta_e\wedge \varphi))\\
		Q3 &= (V_{Q'},\ E_{Q'}\cup\{n^-\overset {e'}{\longrightarrow} m^+\},\ \theta_{Q'}\cup(e',\theta_e\wedge \varphi))\\
		Q4 &= (V_{Q'},\ E_{Q'}\cup\{n^-\overset {e'}{\longrightarrow} m^-\},\ \theta_{Q'}\cup(e',\theta_e\wedge \varphi))
		\end{align*}
	\item \textbf{Añadir predicado a nodo positivo con entorno positivo}: si $n\in V_Q^+$, $\mathcal{N}_Q(n)\subseteq V_Q^+$, y $\varphi\in FORM$, entonces $Q+\{n\wedge \varphi\}$:
		$$\{Q_{\sigma}=(V_{Q'},E_{Q'},\alpha_{Q'}\cup \sigma,\theta_{Q'}\cup(n',\theta_n\wedge\varphi))\ :\ \sigma\in \{+,-\}^{\mathcal{N}_Q(n)}\}$$
	donde $Q'=Cl_Q^{\mathcal{N}_Q(n)}$, y $\{+,-\}^{\mathcal{N}_Q(n)}$ es el conjunto todas las posibles asignaciones de signo a los elementos de $\mathcal{N}_Q(n)$ (el entorno en $Q$ del nodo $n$).
\end{itemize}

\begin{figure}[h]
    \begin{center}
        \includegraphics[scale=0.25]{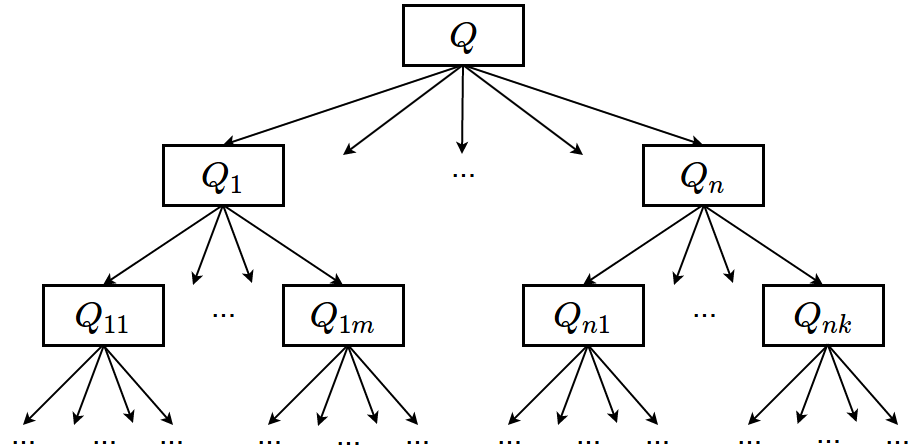}
    \end{center}
    \caption{%
        Árbol de refinamientos.
    }%
    \label{arbolPQG}
\end{figure}

A partir de la estructura de un GGQ no es fácil obtener un GGQ complementario con él. Sin embargo, hay muchos procesos de análisis sobre grafos con propiedades en los que necesitamos trabajar con sucesiones de consultas que verifiquen algunas propiedades de contención y complementariedad como predicados. Los refinamientos vistos en esta sección vienen a cubrir esta carencia y permiten, por ejemplo, construir un árbol de particiones encajadas con los nodos etiquetados de la siguiente forma (Fig. \ref{arbolPQG}):
\begin{itemize}
    \item El nodo raíz está etiquetado con $Q_0$ (un GGQ inicial cualquiera).
    \item Si un nodo del árbol está etiquetado con $Q$, y $R=(Q_1,\dots,Q_n)$ es un conjunto de refinamiento de $Q$, entonces sus nodos hijo se etiquetan con los elementos de $R$.
\end{itemize}

Obsérvese que la construcción del árbol anterior depende por completo de la elección del conjunto de refinamiento que se elija en cada ramificación. 

\section{GGQ-ID3}

GGQ-ID3 es una adaptación del algoritmo ID3 para crear árboles de decisión capaces de clasificar correctamente estructuras inmersas en un Grafo con Propiedades haciendo uso de Generalized Graph Queries como herramientas de test en cada nodo interno del árbol de decisión. 

De forma similar a como trabaja cualquier algoritmo de tipo ID3, nuestra propuesta buscará las consultas que mejor clasifiquen los ejemplos del conjunto de entrenamiento obteniendo los GGQ a evaluar a lo largo de la construcción del árbol (extracción de características) por medio de refinamientos. De esta forma, en cada nodo interno del árbol se realizará una discriminación entre las estructuras que cumplen cada GGQ resultante de un refinamiento. Como suele ser habitual, tanto la medida de impureza usada como la estrategia que elige la mejor ampliación del patrón asociado a un nodo interno es un hiper-parámetro del algoritmo.

Aunque GGQ-ID3 posee una estructura muy similar a los diferentes algoritmos de construcción de árboles de decisión, presenta la novedad de recibir como conjunto de entrenamiento estructuras (subgrafos) inmersas en un grafo con propiedades. Por medio del uso adecuado de predicados, los GGQ del árbol resultante del algoritmo podrán no solo evaluar propiedades de la estructura a clasificar, sino también propiedades de cualquier elemento (subgrafo) en su entorno.

El conjunto de entrenamiento completo, $\mathcal{L}$, estará formado por pares de la forma $(S, valor)$, donde $S$ es un subgrafo de un grafo con propiedades $G$. En consecuencia, cada nodo, $n$, del árbol de decisión construido tendrá asociadas las siguientes estructuras:

\begin{itemize}
\item $\mathcal{L}_n = \{(S_1,y_1),...,(S_N,y_N)\} \subseteq \mathcal{L}$, un subconjunto del conjunto de entrenamiento.
\item $Q_n = (V_{Q_n},E_{Q_n},\alpha_{Q_n},\theta_{Q_n})$, un GGQ que verifican todos los subgrafos de $\mathcal{L}_n$. 
\end{itemize}

El algoritmo \ref{alg:graph-id3} presenta formalmente el algoritmo GGQ-ID3. Nótese que el conjunto de refinamientos disponibles, $REFS$, para ampliar el GGQ	 en cada paso, así como la condición de parada y el criterio de selección del refinamiento, permanecen como parámetros libres del modelo.

\begin{algorithm}
\begin{algorithmic}[1]
\STATE {Create a $Root$ node for the tree}
\IF {Stop criteria is reached}
\RETURN The single-node tree $Root$, with most frequent label in $\mathcal{L}$.\label{op0}
\ELSE
\STATE {$(Q_1,..,Q_k) = Optimal\_Refinement(G,Q,L,REFS)$}
\STATE {$\mathcal{L}_1= \{(S,y) \in \mathcal{L} : S\vDash Q_1\},...,\mathcal{L}_k= \{(S,y) \in \mathcal{L} : S\vDash Q_k\}$}
\STATE {Add $k$ new tree branches below $Root$ with values GGQ-ID3($G,Q_i,\mathcal{L}_i,REFS$) for every $1\leq i \leq k$.}
\ENDIF
\end{algorithmic}
\caption{GGQ-ID3($G,Q,\mathcal{L},REFS$)}\label{alg:graph-id3}
\end{algorithm}

En un proceso habitual de aprendizaje a partir de subgrafos de $G$, los parámetros de la llamada inicial del algoritmo serán:

\begin{itemize}

\item $G = (V,E,\mu)$, grafo en el que se encuentran inmersos las estructuras a clasificar.

\item $\mathcal{L}= \{(S_1,y_1),...,(S_N,y_N)\}$, conjunto de pares $(S_i,y_i)$ donde $S_i \subseteq G$ representa un subgrafo e $y_i$ es su valor de salida asociado. 

\item $Q_0 = (V_Q,E_Q, \mu_Q)$, GGQ inicial (normalmente un GGQ con la estructura común más grande en $S_1,...,S_N$).

\item $REFS$, conjunto de refinamientos disponibles.

\end{itemize}

El algoritmo sigue el procedimiento habitual en un algoritmo de tipo ID3. Comienza creando un árbol que contiene un único nodo (nodo raíz) al que están asociados todos los objetos de $\mathcal{L}$ y $Q_0$. Si $n$ es el nodo del árbol de decisión con el que estamos trabajando, el algoritmo evalúa qué refinamiento permite dividir de mejor manera $\mathcal{L}_n$ (máxima reducción de impureza) y éste será elegido para ser aplicado a $Q_n$. A continuación se crearán tantas ramas a partir de $n$ como GGQs tenga el refinamiento elegido y por cada una de ellas se transmitirán los pares de $\mathcal{L}_n$ que cumplan con el GGQ asociado. Cada nodo hijo de $n$ heredará cada GGQ resultante del refinamiento y procederá a buscar (si no se alcanza la condición de parada) el mejor refinamiento para el nuevo GGQ. De esta manera, por cada rama del árbol se heredará, no sólo un conjunto de pares, sino un GGQ que verifican todos ellos. En caso de que se cumpla la condición de parada, el nodo se convertirá en un nodo hoja asociada a la clase correspondiente. Este proceso se repite de manera recursiva.

Los árboles de decisión que se obtienen en la mayoría de los procedimientos automáticos dividen los datos en subconjuntos complementarios de manera recursiva \cite{c45}. Habitualmente, la división se lleva a cabo mediante la evaluación de una condición sobre el conjunto de objetos actual y su correspondiente división binaria: una rama recibirá el conjunto de objetos que cumplen con la condición, mientras que la otra recibe aquellos que no la cumplen. En los casos simples los patrones complementarios binarios se pueden producir simplemente negando la condición, pero en el caso multi-relacional la producción de patrones complementarios de este tipo no es tan directa. Es por ello que en la sección \ref{PQG} se presentaron una serie de refinamientos que generan GGQ complementarios, aunque no necesariamente binarios. Nosotros usaremos esos refinamientos como conjunto $REFS$ en la llamada del algoritmo en los ejemplos que se mostrarán a continuación, pero podría enriquecerse con cualquier refinamiento adicional que se considerase de valor.

\subsection{Ejemplo de aplicación del algoritmo GGQ-ID3}

Vamos a presentar un caso concreto de aplicación del algoritmo GGQ-ID3 sobre un pequeño grafo con propiedades a modo de demostración. Los refinamientos serán los vistos en la sección \ref{PQG},  usaremos la condición de parada más restrictiva (que todos los pares del nodo actual pertenezcan a la misma clase), y se usará la Ganancia de Información \cite{Mitchell:1997:ML:541177} como medida de impureza.

Trabajaremos con el grafo social mostrado en la Figura \ref{grafo1}, en el que se representan algunas conexiones de tipo marital entre usuarios, y otras relacionadas con la publicación de fotografías por estos usuarios. Podemos encontrar nodos de tipo \texttt{user} y \texttt{photo}, y relaciones de tipo \texttt{husband}, \texttt{wife}, \texttt{publish} y \texttt{likes}. En este caso, no potenciaremos $L$ por medio de medidas topológicas del grafo, por lo que en el caso del refinamiento \textbf{añadir predicado a nodo} los predicados disponibles serán $\{type=photo,type=user\}$, y en el caso del refinamiento \textbf{añadir predicado a arista}, los predicados disponibles serán $\{type=publish,type=likes,type=husband\}$.

\begin{figure}[h!]
  \centering
  \includegraphics[width=\textwidth]{png/arboles/grafo1.png}
  \caption{Grafo Social.}
  \label{grafo1}
\end{figure}

Adicionalmente, tenemos una propiedad en algunos nodos (los de tipo \texttt{user}) que indica su género, con posibles valores \texttt{F} (female) y \texttt{M} (male), los nodos de tipo \texttt{photo} tendrán asociado un valor de género \texttt{None}. 

Aunque como hemos indicado el algoritmo GGQ-ID3 está preparado para construir árboles de decisión capaces de clasificar cualquier estructura (subgrafo) en un grafo con propiedades, con el fin de mostrar un primer ejemplo simple, abordamos un problema que solo intenta clasificar nodos, e intentaremos predecir el valor del género en los mismos. Con el objetivo de que la exposición resulte más clara, y para evitar confusiones con la terminología, a los nodos pertenecientes al conjunto de entrenamiento los denominaremos \emph{objetos}, mientras que dejaremos el término \emph{nodos} para los nodos del árbol de decisión en construcción.

\begin{figure}[h!]
	\centering
	\includegraphics[scale=0.25]{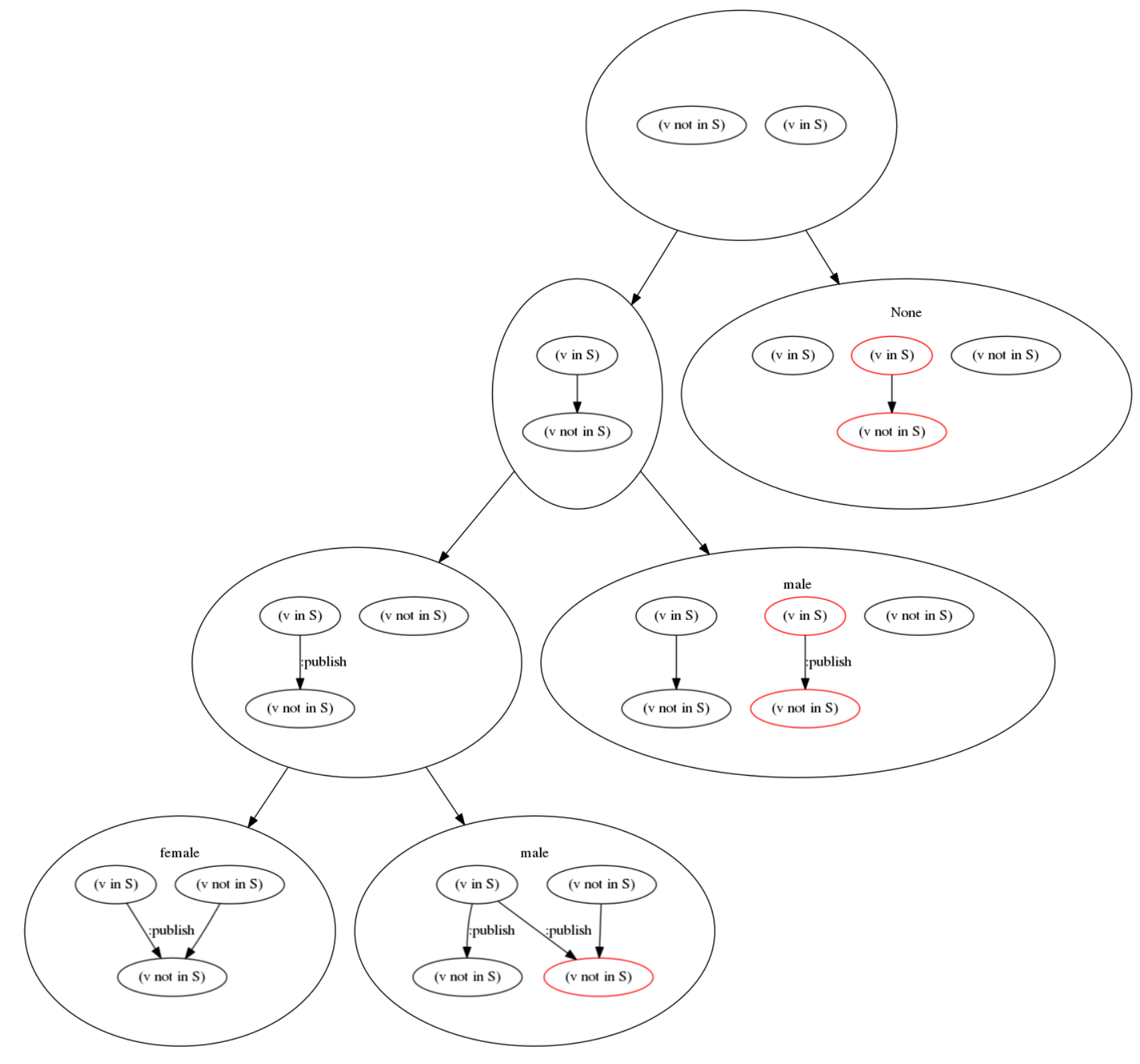}
	\caption{Árbol de decisión GGQ (grafo social).}
	\label{social1}
\end{figure}

El algoritmo GGQ-ID3 construye el árbol de decisión de la Figura \ref{social1} (los nodos/aristas positivas son marcados en negro y los negativos en rojo) que es capaz de clasificar correctamente todos los nodos del grafo social de la Figura \ref{grafo1} según su género: \texttt{F}, \texttt{M} o \texttt{None}. Todas las indicaciones que se hagan acerca de la construcción siguen la representación mostrada en la citada figura.

El conjunto de entrenamiento inicial, $\mathcal{L}$, lo forman todos los pares formados por objetos del grafo de datos y su correspondiente género. En la ejecución del algoritmo el GGQ inicial, $Q_0$, está compuesto por dos nodos positivos, uno de ellos con un predicado que exige pertenencia al subgrafo bajo evaluación ($v \in S$) y otro que exige la condición contraria ($v \notin S$). Como se ha comentado anteriormente, es habitual que el GGQ inicial corresponda a la mayor subestructura común en los subgrafos a clasificar, en este caso, la mayor estructura común está compuesta por un único nodo sin restricciones. Además, y por motivos de eficiencia, en cada paso del algoritmo se creará un nodo positivo aislado con un predicado que exige la no pertenencia al subgrafo bajo evalacuión ($v \notin S$) si es que no existe ningún nodo aislado de este tipo en el GGQ actual. Si el subgrafo bajo evaluación no cubre todos los nodos del grafo en el que se encuentra inmerso (lo cual se cumple en todos los ejemplos presentados), añadir un nodo de este tipo no modifica la semántica asociada al GGQ al que es añadido.
 
Como primer paso, el algoritmo analizará qué refinamiento de $REFS$ (y con qué parámetros) aporta mayor ganancia de información, dando como resultado el refinamiento $+\{\overset {e^+}{\longrightarrow}\}$ entre los únicos dos nodos existentes en $Q_0$. Por lo que el primer nodo (raíz) del árbol de decisión será el que realice el test sobre la existencia o no de una arista saliente en alguno de los nodos en el subgrafo a evaluar. Aunque el refinamiento \textbf{añadir arista} genera un refinamiento de cuatro elementos, consideramos solo aquellos que intervienen en el proceso de clasificación, y no los que transmiten un conjunto vacío de objetos.

Debido a que los subgrafos que se quieren clasificar están compuestos por un único objeto, este refinamiento evalúa la existencia o no de una arista saliente en este objeto. Las ramas correspondientes a la existencia (respectivamente, no existencia) de dicha relación trasmitirán los objetos que posean (respectivamente, no posean) una relación saliente. De acuerdo a la representación dada en la Figura \ref{social1}, por la rama derecha se trasmitirán todos los objetos que no posean una relación saliente (en este caso concreto, todos los objetos del grafo de datos de tipo \texttt{photo} y ninguno de tipo \texttt{user}), ninguno de estos objetos poseen valor asociado a la propiedad \texttt{genre} por lo que directamente esta rama queda asociada al valor de salida \texttt{None} y genera una hoja del árbol de decisión (ya que presenta pureza máxima respecto de la clasificación buscada). Por la rama izquierda se trasmitirán todos los objetos que posean una relación saliente (en este caso, todos los objetos del grafo de datos de tipo \texttt{user} y ninguno de tipo \texttt{photo}). 

Como los valores de la propiedad \texttt{genre} no son homogéneos para estos objetos (este nodo del árbol de decisión actual presenta impureza), el algoritmo no acaba y es necesario aplicar un nuevo refinamiento a esta rama que produzca alguna ganancia de información adicional. Recordemos que hasta este nodo han llegado los objetos que reflejan usuarios masculinos y femeninos con una arista saliente. De nuevo, se debe evaluar qué refinamiento aporta una mayor ganancia de información. 

El refinamiento $+\{\overset {e \wedge \{type=publish\}}{\longrightarrow} \}$, \textbf{añadir un nuevo predicado a la arista saliente}, es el que mayor ganancia de información aporta. De nuevo, aunque este refinamiento contiene cuatro GGQ, representamos solo aquellos que intervienen en el proceso de clasificación.

Los refinamientos aplicados a este nodo del árbol de decisión discriminarán qué objetos (de los que le llegan, que son los que tienen una arista saliente) verifican que esta arista es de tipo \texttt{publish} y cuáles no. De acuerdo a la representación dada en la Figura \ref{social1}, por la rama derecha de dicho nodo se transmitirán los objetos que no tengan una arista saliente de tipo \texttt{publish}. En el caso que estamos clasificando, todos estos objetos son usuarios de género masculino, por lo que se alcanza la condición de parada de máxima pureza y el nodo pasa a ser una hoja del árbol de decisión asociada a la clase \texttt{M}. Por la rama izquierda del nodo se transmitirán los objetos del grafo de datos que tengan una arista saliente de tipo \texttt{publish}. En este caso, de nuevo los valores de la propiedad \texttt{genre} no son homogéneos (presentan impureza) por lo que el algoritmo debe continuar buscando un nuevo refinamiento en esta rama.

Procedemos, pues, a repetir el procedimiento para este nodo, al que han llegado los objetos que se corresponden con usuarios con propiedades \texttt{M} y \texttt{F} con una arista saliente de tipo \texttt{publish}. El refinamiento que mayor ganancia de información aporta es $+\{\overset {e^+}{\longrightarrow}\}$, \textbf{añadir una arista} entre el nodo aislado ($v \notin S$) y el nodo destino de la relación tipo \texttt{publish}. De nuevo, tomamos en cuenta solo aquellos refinamientos por los que se transmiten objetos del grafo de datos original. La interpretación a este nivel del árbol de decisión es discriminar, de entre los objetos que han publicado algo, aquellos cuya publicación recibe alguna arista entrante por otro objeto que no pertenece a la estructura bajo evaluación y los que no. 

De las dos ramas que producen algún tipo de filtrado efectivo, por la derecha (siempre según la representación de la Figura \ref{social1}) se transmitirán los objetos que no cumplan con dicha condición. Todos estos objetos son usuarios de género \texttt{M} (y representan usuarios que han publicado una foto que no le gusta a nadie), por lo que se alcanza la condición de parada (máxima pureza) y el nodo se convierte en una hoja del árbol de decisión asociada a la clase \texttt{M}. Por la rama izquierda se transmitirán los usuarios que hayan publicado una fotografía y les haya gustado a alguien (esta fotografía publicada tiene una arista entrante que no proviene del nodo bajo evaluación). Todos estos usuarios son de género \texttt{F}, por lo que se verifica también la condición de parada y la hoja producida queda asociada a la clase correspondiente.

De esta manera, obtenemos un árbol de decisión que es capaz de clasificar correctamente todos los objetos en el grafo de datos asignándolos correctamente a la clase a la que pertenecen según su género haciendo uso de la estructura multi-relacional en la que se encuentran inmersos. 

Además, la interpretación de los diversos nodos del árbol de decisión muestra claramente cómo, por medio de los Generalized Graph Queries y el sistema de refinamientos construido, se pueden conseguir refinamientos que evalúan propiedades del contexto diferenciando entre los objetos que deben estar dentro de la estructura analizada y aquellos que deben estar fuera, ampliando considerablemente la capacidad expresiva del sistema de consulta y, en consecuencia, su capacidad discriminadora.

\section{Ejemplos de Aplicación}

A continuación presentamos algunos ejemplos de árboles de decisión multi-relacio\-na\-les que hacen uso de GGQ y que han sido obtenidos siguiendo el algoritmo GGQ-ID3 presentado. Al igual que en el caso explicado anteriormente, se han ido añadiendo nodos positivos aislados no pertenecientes al subgrafo bajo evaluación en cada paso si no existían en el GGQ y se ha partido de GGQ iniciales que contenían un nodo positivo con un predicado que lo fija al subgrafo bajo evaluación y otro nodo positivo con la restricción contraria. 

Los ejemplos han sido extraídos de bases de datos en grafo pequeñas pero con la suficiente complejidad como para mostrar la capacidad de descubrimiento de patrones que posee el algoritmo GGQ-ID3. No mostraremos los árboles completos resultantes (debido a la falta de resolución del papel para ser mostrados adecuadamente) sino sólo las hojas clasificadoras o algunas de las ramas más interesantes.

\subsection{StarWars}

\begin{figure}[htb]
    \begin{center}
        \includegraphics[width=\textwidth]{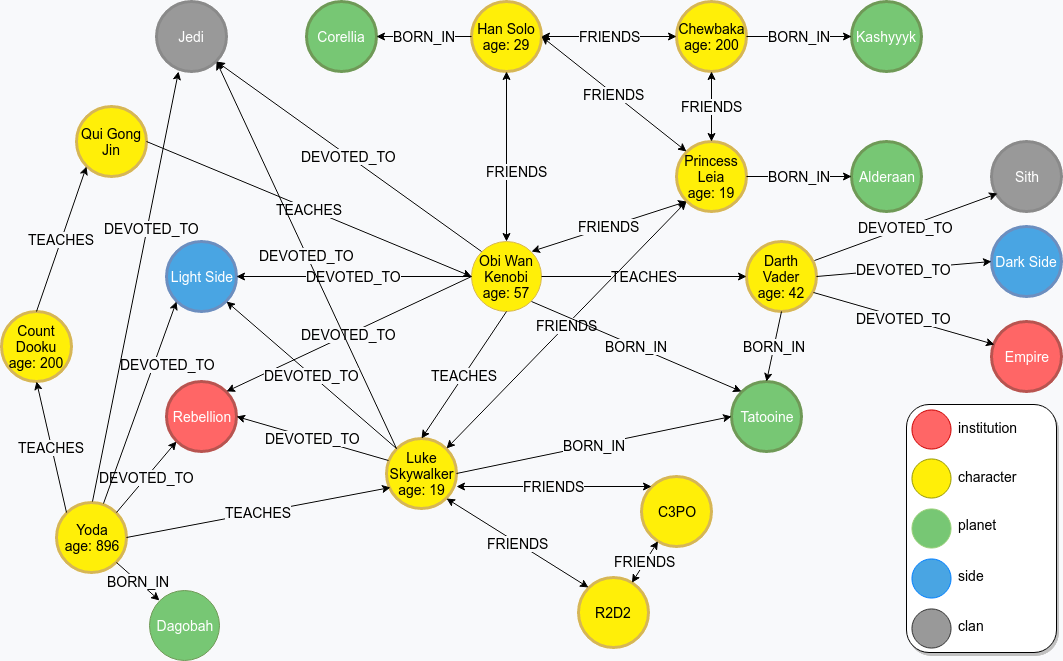}
    \end{center}
    \caption{%
        Grafo Starwars.
    }%
    \label{starwars}
\end{figure}

El primer ejemplo obtenido a través del algoritmo GGQ-ID3 lo conseguimos minando el grafo presentado en la Figura \ref{starwars} con información sobre StarWars \footnote{http://console.neo4j.org/?id=StarWars}.

\begin{figure}[htb]
  \centering
  \includegraphics[width=\textwidth]{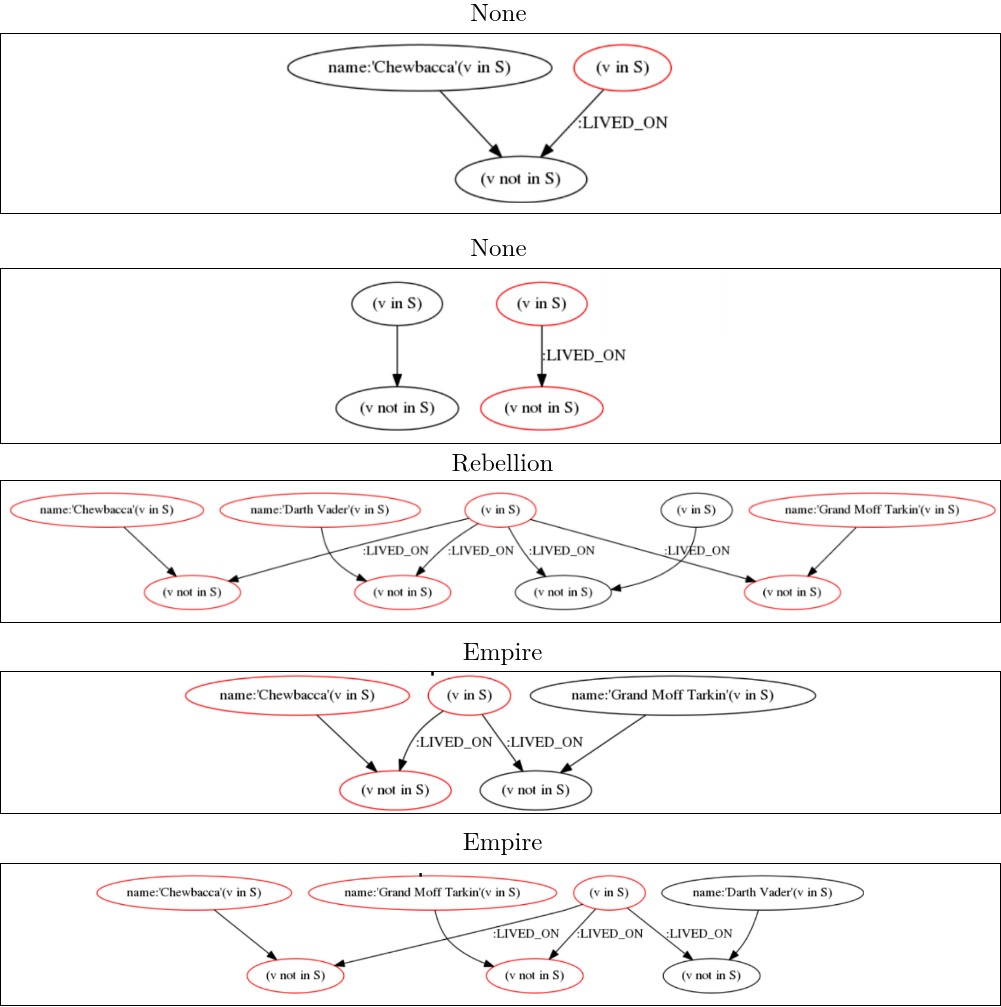}
  \caption{Hojas del árbol de decisión GGQ (grafo Starwars)}
  \label{starwars1}
\end{figure}

Los GGQ presentados en la Figura \ref{starwars1} permiten discriminar cada personaje presente en el grafo según si es devoto del Imperio, de la Rebelión o de ninguno de los dos bandos. Se corresponden con las diversas hojas clasificadoras del árbol de decisión calculado automáticamente. Para ello, los nodos de tipo \texttt{institution} (junto con las aristas en las que participan) han sido eliminados de dicho grafo. Las posibles clases en las que clasifica el árbol obtenido son: \texttt{empire}, \texttt{rebellion} o \texttt{None} (en el caso en el que el personaje no sea devoto de ninguna de las dos instituciones).

Los GGQ obtenidos en las hojas muestran que incluso trabajando con grafos relativamente pequeños el nivel de complejidad que pueden alcanzar las consultas proporcionan ejemplos muy interesantes de aprendizaje de patrones de forma automática usando la metodología presentada.

\subsection{El Hobbit}

\begin{figure}[htb]
  \centering
  \includegraphics[width=\textwidth]{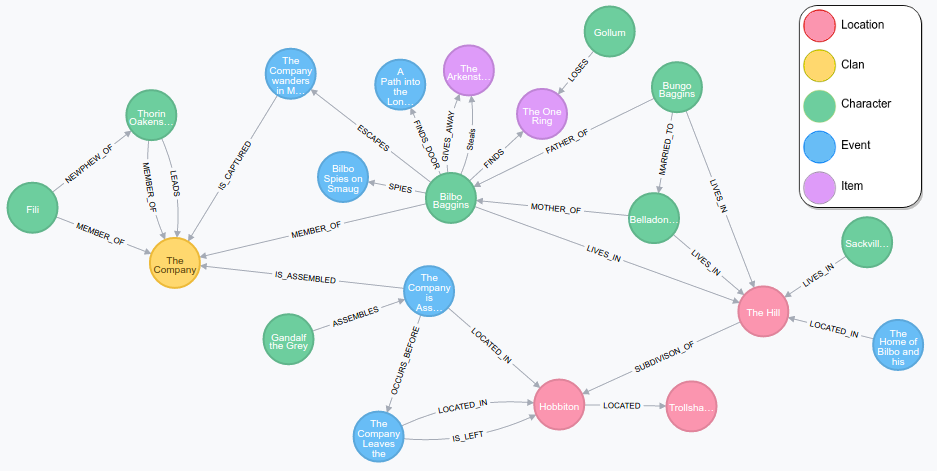}
  \caption{Sección del grafo El Hobbit.}
  \label{ejhobbit}
\end{figure}

El segundo ejemplo lo obtenemos minando otro grafo de juguete habitual en las pruebas realizadas para bases de datos en grafo, en este caso relacionado con la historia del Señor de los Anillos \footnote{\url{http://neo4j.com/graphgist/c43ade7d259a77fe49a8}}. Este grafo contiene 105 nodos distribuidos a través de 7 tipos (\texttt{Character}, \texttt{Event}, \texttt{Item}, \texttt{Clan}, \texttt{Aligment}, \texttt{Location} y \texttt{Chapter}) y 209 aristas distribuidas a través de 65 tipos. Parte de su interés para hacer pruebas de aprendizaje de estructuras radica en que presenta una tipología en aristas muy elevada, con muy pocos representantes de algunos tipos de aristas, por lo que mecanismos ineficientes tenderán a crear árboles muy grandes como único método para poder realizar clasificaciones multi-relacionales. La Figura \ref{ejhobbit} muestra un subgrafo extraído de esta pequeña base de datos.

\begin{figure}[htb]
  \centering
  \includegraphics[width=\textwidth]{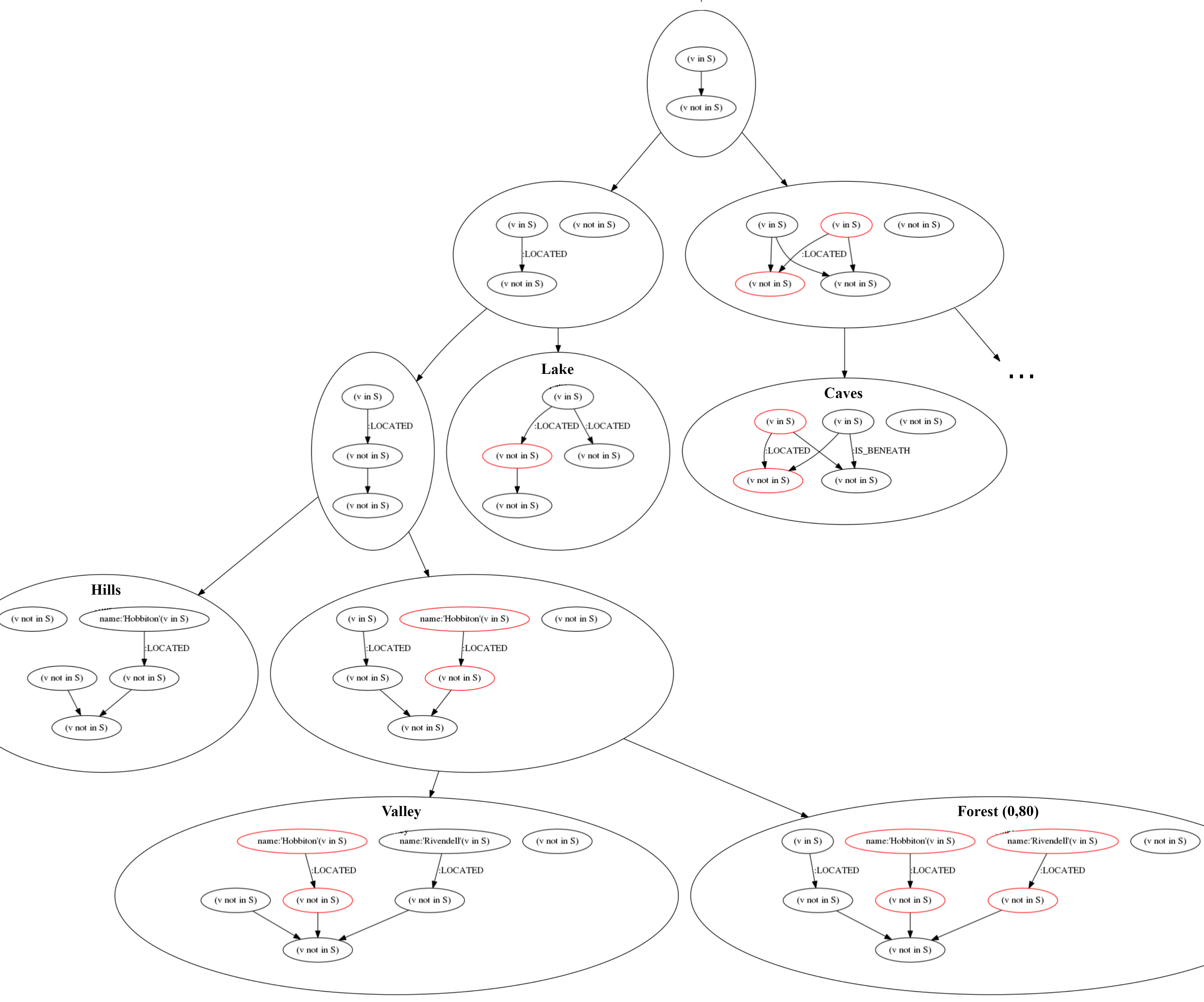}
  \caption{Árbol de decisión GGQ (grafo El Hobbit).}
  \label{hobbit1}
\end{figure}

El árbol de decisión presentado en la Figura \ref{hobbit1} se ha obtenido automáticamente por medio de GGQ-ID3 y permite discriminar los posibles tipos de ubicación (\texttt{Loca\-tion}) presentes (\texttt{Hills}, \texttt{Forest}, \texttt{Valley}, \texttt{Mountain}, \texttt{Caves} y \texttt{Lake}). En la construcción del árbol se ha impuesto una profundidad máxima de 5 niveles y se han eliminado algunas ramas por motivos de presentación.

\section{Algunas Notas sobre la Implementación}

Con el fin de hacer pruebas de verificación sobre el sistema de consultas creado (GGQ) y el algoritmo GGQ-ID3 que hace uso del mismo, se ha llevado a cabo una implementación como prueba de concepto \footnote{\url{https://github.com/palmagro/ggq}} en el lenguaje de programación Python que hace uso de la base de datos en grafo Neo4j \footnote{\url{http://neo4j.org}} como sistema de persistencia. La verificación de un GGQ sobre subgrafos almacenados en Neo4j ha sido realizada apoyándose en el lenguaje de consultas Cypher, donde las evaluaciones relacionadas con la existencia de caminos son muy expresivas y están altamente optimizadas. 

Sin lugar a dudas, esta implementación ganaría en eficiencia si, en lugar de haber desarrollado el sistema de consulta sobre el lenguaje Cypher, se hubiera utilizado la API Java de Neo4j o incluso implementando un sistema de persistencia \textit{ad hoc} orientado a soportar este tipo de tareas, pero nuestro objetivo ha sido el de demostrar que el sistema formal de consultas es capaz de realizar este tipo de tareas de forma sencilla.

En la implementación se han realizado algunas optimizaciones leves, pero que son determinantes para poder llevar a cabo la tarea de forma efectiva, como por ejemplo aprovechar la complementariedad de los GGQ resultantes de un refinamiento para ahorrar tiempo en las consultas y podar directamente las hojas a las que no llega ningún elemento del conjunto de entrenamiento. 

Por la forma en que se construyen los refinamientos, y como se almacenan en el árbol de decisión, si un objeto verifica un GGQ, también verificará todos los GGQ antecesores de éste, por lo que la potencia clasificadora de un árbol de decisión obtenido a través del algoritmo GGQ-ID3 está contenida en las hojas del árbol. Sin embargo, para ganar eficiencia, a la hora de clasificar un ejemplo nuevo a partir del árbol obtenido, se aconseja usar el árbol completo, ya que solo serán necesarias, a lo sumo, $k$ evaluaciones (con $k$ la profundidad del árbol), pero puede haber una cantidad exponencial de hojas clasificadoras. Además, como de un nivel al siguiente se ha producido una modificación atómica, aportada por el refinamiento correspondiente, la evaluación de un subgrafo en un nodo del árbol de decisión supone considerar únicamente algunas comprobaciones adicionales a la evaluación de su nodo padre.

\section{Conclusiones y Trabajo Futuro}

En este artículo se ha presentado el algoritmo GGQ-ID3, que hace uso de los Generalized Graph Queries como herramientas de test para la construcción de un árbol de decisión siguiendo los fundamentos del algoritmo ID3. En los resultados de los experimentos llevados a cabo, se muestra que GGQ-ID3 es capaz de extraer patrones interesantes que pueden ser utilizados en tareas de aprendizaje complejas. Para ello, basta considerar los GGQ contenidos en las hojas como nuevos atributos descubiertos por el algoritmo. De esta forma, además de construir un árbol clasificador, el algoritmo es capaz de descubrir patrones que caracterizan diferentes estructuras en el grafo (\textit{graph pattern mining}) y que pueden ser utilizados como atributos de las estructuras clasificadoras en tareas posteriores (\textit{feature extraction}).

Hemos mostrado un conjunto inicial de refinamientos a la hora de presentar ejemplos de aplicación del algoritmo GGQ-ID3, pero este conjunto puede ser modificado añadiendo refinamientos acordes a la estructura del grafo utilizado en el aprendizaje o a la tarea a la que se orienta el mismo.

El algoritmo MRDTL fue desarrollado hace más de una década para trabajar específicamente en bases de datos relacionales y con tareas simples de clasificación, y puede ser visto como un caso particular del algoritmo GGQ-ID3 en el que sólo se permiten GGQ con forma de árbol (ya que hacen uso de grafos de selección) y donde sólo se permite aprendizaje a partir de estructuras formadas por un único nodo. En este sentido, GGQ-ID3 supone un salto adelante en una línea de trabajo iniciada hace años y que se consideraba estancada desde entonces. 

El principal problema que presentan los algoritmos de construcción de árboles de decisión multi-relacionales es que el espacio de hipótesis es extremadamente grande, y evidentemente GGQ-ID3 no está libre de este problema. Para reducir su complejidad y orientar la búsqueda se pueden proponer varias soluciones. Por un lado, se puede analizar de manera estadística la frecuencia de aparición de ciertas estructuras atendiendo a las propiedades que intervienen y a las restricciones asociadas con el fin de reducir el número de posibles refinamientos a aplicar en cada caso y reducir el coste de la búsqueda del mejor refinamiento. Para ello es necesario hacer uso de diversas medidas desarrolladas para grafos generalizados que extienden las medidas de frecuencia más simples que se usan en el caso de MRDTL-2 \cite{Atramentov2003}. Por otro lado, se pueden crear familias de refinamientos más complejos (por ejemplo, combinar el refinamiento \textbf{añadir arista} con \textbf{añadir propiedad a una arista} en un solo paso) para de esta manera reducir el número de pasos para obtener GGQ complejos y ampliar la reducción de impureza que suponen los pasos atómicos que son menos informativos. Si se lleva a cabo esta última opción de manera adecuada (unificando los refinamientos en función de la frecuencia de aparición de estructuras en el grafo) se puede conseguir que el algoritmo se acerque de manera más rápida a la solución. En ambos casos se consigue una mejora en la eficiencia sacrificando la posibilidad de cubrir un espacio de hipótesis más amplio (pero que probablemente ofrece alternativas en las que la reducción de impureza es menor). En este sentido, en este trabajo se ha ofrecido un conjunto minimal de refinamientos bien construidos, pero debe tenerse en cuenta que no se ofrecen con la intención de que sea óptimo para ciertas tareas de aprendizaje.

El segundo gran problema que tiene el agoritmo GGQ-ID3 (y que es heredado por todos los algoritmos inspirados en ID3) es la imposibilidad de deshacer las decisiones tomadas durante la construcción del árbol. De tal manera que las opciones de refinamiento en un paso determinado del algoritmo dependen de los refinamientos elegidos en pasos anteriores y determinan, hasta cierto punto, las opciones futuras. Para solucionar este problema, es habitual utilizar algún procedimiento de \textit{backtracking} que permita deshacer decisiones si han desembocado en un mal resultado, o algún procedimiento de \textit{Beam-Search}, como el utilizado en el algoritmo GBI \cite{Geamsakul}, que permita tomar varias decisiones en paralelo, y finalmente seleccionar la que haya derivado en una mejor solución.

Con respecto a los trabajos futuros que derivan del desarrollo aquí presentado, cabe mencionar que, gracias a que los GGQ están construidos utilizando la estructura de grafo generalizado, y que dicha estructura permite la definición de hipergrafos de manera natural, los GGQ pueden evaluar hipergrafos con propiedades teniendo en cuenta pequeñas modificaciones sobre las definiciones presentadas, por lo que la extensión de los Generalized Graph Queries hacia Generalized Hypergraph Queries y por tanto a un GGQ-ID3 capaz de aprender de hipergrafos es un paso natural que merece la pena ser considerado. Además, el desarrollo de diferentes conjuntos de refinamiento en función del tipo de grafo a consultar o incluso la generación automática de dichos conjuntos a partir de estadísticas extraídas del grafo a analizar puede derivar en optimizaciones importantes en procesos de construcción automática y efectiva de GGQ. Por útlimo, cabe destacar que a pesar de que los GGQ ya están siendo utilizados por procedimientos de descubrimiento/aprendizaje como el algoritmo GGQ-ID3, son grandes candidatos para ser utilizados por otros algoritmos de este tipo. 

Los árboles de decisión constituyen una herramienta idónea de Aprendizaje Automático para ser combinada a través de algún modelo \emph{ensemble}, como Random Forest, gracias a su bajo coste computacional en el entrenamiento y a la aletoriedad conseguida en el modelo a partir de pequeños cambios en el conjunto de datos de los que aprender. Por ello, disponer de modelos adecuados de generación automática de árboles de decisión multi-relacionales se convierte en una tarea de gran importancia en el conjunto de la Inteligencia Artificial.

El aprendizaje automático que hace uso de información relacional ha estado (y sigue estando) en un segundo plano en relación al aprendizaje automático más estándar, que hace uso de información no relacional, habitualmente en forma de tablas y otras estructuras más regulares. Las bases de datos que más habitualmente se han utilizado, y en las que se encuentra almacenada la información referente a la mayoría de los fenómenos estudiados, hacen uso de esquemas y sistemas basados en bases de datos relacionales, que no muestran un desempeño óptimo al trabajar con relaciones complejas. Además, la mayor riqueza expresiva de las estructuras de representación de la información más complejas impone una mayor dificultad a la hora de realizar nuevos algoritmos y proporciona, al menos en las primeras aproximaciones, resultados menos llamativos que los métodos más depurados y más tradicionales.

Con respecto al aprendizaje en grafos con propiedades, cabe destacar que existen varias líneas de trabajo que transforman los datos originales (en forma de grafo) hacia otras estructuras que los algoritmos más extendidos son capaces de manejar de manera más natural (debido a que fueron creados para trabajar específicamente con dichas estructuras). Es el caso de las inmersiones de grafos en espacios vectoriales \cite{deepwalk,Tang:2015:LLI:2736277.2741093}, así como de la propuesta presentada en \cite{grover2016node2vec}, que muestrean el grafo a partir de subestructuras para acomodarlo a una colección de objetos (pares $(elemento,contexto)$) que los algoritmos tradicionales pueden consumir de manera óptima. También es el caso de los trabajos que usan Redes Neuronales Convolucionales para realizar tareas de aprendizaje en grafos \cite{kipf2016semi,DBLP:journals/corr/DefferrardBV16,NIPS2015:5954}. Para poder utilizar este tipo de modelos sobre grafos debemos definir qué se entiende por el contexto espacial de un elemento del grafo, haciendo suposiciones que incluyen un sesgo adicional a la información analizada. Desde nuestro punto de vista, éstas son aproximaciones válidas que deben seguir siendo investigadas, pero se deben considerar otras opciones como la de trabajar directamente con la estructura de grafo, que ha sido una de las líneas de investigación seguidas en este trabajo y que ha demostrado ser válida.

Otra caractertística importante de los métodos de aprendizaje basados en árboles de decisión es que representan un modelo de caja blanca, ofreciendo una explicación interpretable por un humano de las decisiones tomadas a la hora de realizar regresión o clasificación. En el caso del algoritmo GGQ-ID3, este característica es potenciada gracias a la interpretabilidad de los GGQ. 

El hecho de que perdamos capacidad en la interpretación de un resultado al combinar árboles multi-relacionales no impide que algoritmos como GGQ-ID3 puedan ser combinados en este tipo de métodos agregados para llevar a cabo tareas de predicción de tipo caja negra. Sin embargo, existen posibilidades para combinar diferentes árboles de este tipo y que sigan ofreciendo justificaciones interpretables por los humanos. Una posibilidad es combinar los GGQ de hojas asociadas a la misma clase en los diferentes árboles, dando lugar a patrones combinados (posiblemente de manera probabilística) que son capaces de condensar los diferentes predicados que caracterizan a una misma clase en un predicado más potente.

\bibliographystyle{plain}	
\bibliography{biblio}